\documentclass{ws-ijitdm}

\usepackage{placeins}
\usepackage[english]{babel}
\usepackage[square,numbers]{natbib}
\usepackage{url}

\usepackage{tabularx} 
\usepackage{booktabs}
\usepackage{hyperref}
\hypersetup{colorlinks=true, linkcolor=blue, filecolor=magenta, urlcolor=red, pdfpagemode=FullScreen,}

\begin{document}
\renewcommand{\footnotesize}{\fontsize{9pt}{11pt}\selectfont}
\title{Linear Programming for Multi-Criteria Assessment with Cardinal and Ordinal Data: A Pessimistic Virtual Gap Analysis}

\author{Fuh-Hwa Franklin Liu \footnotemark{}\footnotetext{Corresponding author; fliu@nycu.edu.tw} }
\address{Professor Emeritus, Department of Industrial Engineering and Management,\\  National Yang Ming Chiao Tung University, Taiwan 300, Republic of China}
\author{Su-Chuan Shih \footnotemark{}\footnotetext{scshih@gm.pu.edu.tw}}
\address{Associate Professor, Department of Marketing and Digital Business Administration,\\ Providence University, Taiwan 433, Republic of China}

\maketitle

\begin{abstract}
   Multi-criteria Analysis (MCA) is used to rank alternatives based on various criteria. Key MCA methods, such as Multiple Criteria Decision Making (MCDM) methods, estimate parameters for criteria to compute the performance of each alternative. Nonetheless, subjective evaluations and biases frequently influence the reliability of results, while the diversity of data affects the precision of the parameters. The novel linear programming-based Virtual Gap Analysis (VGA) models tackle these issues. This paper outlines a two-step method that integrates two novel VGA models to assess each alternative from a pessimistic perspective, using both quantitative and qualitative criteria, and employing cardinal and ordinal data. Next, prioritize the alternatives to eliminate the least favorable one. The proposed method is dependable and scalable, enabling thorough assessments efficiently and effectively within decision support systems.

\end{abstract}

\section{Objectives of the Research}\label{sec:1}
\subsection {Issues to Multi-Criteria Assessment (MCA)} \label{sec:1.1}

In the past fifty years, numerous notable \textbf{multi-criteria assessment (MCA)} methodologies have been documented. These include \textbf{parametric} methods, multiple-criteria decision-making (MCDM), and stochastic frontier analysis (SFA), as well as\textbf{ semi-parametric} methods like Data Envelopment Analysis (DEA), among other techniques. 

These MCA techniques are utilized across various domains, such as social science, operations management, engineering, and production systems, involving \textit{n} \textbf{alternatives} based on \textbf{minimization and maximization criteria}, all quantified in distinct units. In production settings, MCA methods are employed to evaluate the efficiency of \textit{n} \textbf{decision-making units (DMUs)} using \textit{m} \textbf{inputs} and \textit{s} \textbf{outputs}. The terms alternatives and criteria (such as minimization and maximization) can be synonymously referred to as DMUs and performance indices (such as input and output), respectively. MCA methods are applied to the decision matrix, which has (\textit{m+s}) rows and \textit{n} columns, constructed from the observed data of alternatives assessed against the criteria. 

\citet{Liu2025a} introduced the unique \textbf{non-parametric} method known as the linear programming-based Virtual Gap Analysis (VGA) to address the shortcomings present in parametric methods such as MCDM, SFA, and DEA. The study by \citet[in][Figure 1]{Liu2025b} is modified and illustrated in Figure \ref{fig1}, which shows the types of MCA challenges tackled by VGA-methods through VGA-models. Within each VGA-model, a specific $DMU_o$ is evaluated against others to gauge its performance and to derive a set of parameters from the linear model. Each DMU alternately serves as $DMU_o$.

\subsection{Categorization of VGA-methods and VGA-models}\label{sec:1.2}
In Figure \ref{fig1}, the eight VGA-methods—specifically \textbf{b1c, b2c, w1c, w2c, b1v, b2v, w1v, and w2v}—illustrate combinations of three orientations. Each of these VGA-methods comprises multiple VGA-models to conduct accurate assessments. 

\begin{enumerate} 
\item 
Orientation 2: Best practice vs worst practice (b vs w)

Examining DMUs with an optimistic viewpoint and a pessimistic viewpoint requires different methodologies. In the \textbf{best practice} assessment, each of the \textit{n} DMUs aims to reduce inputs while enhancing outputs to boost performance. Conversely, in the \textbf{worst practice} assessment, increasing inputs and reducing outputs may lead to a decline in performance.

\item 
Orientation 2: Type 1 vs Type 2 decision matrices (1 vs 2)

 \textbf{Type 1} consists solely of cardinal data, while \textbf{Type 2} is a combination of cardinal and ordinal data. Ordinal data is employed to represent unique qualitative attributes through subjective evaluations, utilizing methods such as group consensus, surveys, or voting, and is often illustrated with Likert scales. A Likert scale typically spans M positions, with M ideally being between 3 and 7. A DMU that performs well in output (input) should be placed in the \textit{M-th} (first) position.

 Other varieties of decision matrices might include cardinal data that can have zero or negative values, along with data that are imprecise and bounded. The \textbf{nominal data} metrics are unsuitable for MCA. For instance, metrics such as race, religion, gender, and color consist of distinct categories, rendering any continuous measurement irrelevant in this context.

\item 
Orientation 3: Perspective-c vs Perspective-v (c vs v)

In \textbf{Perspective-c}, the evaluation team uses a dual-stage approach to evaluate DMUs. During Stage I of best (or worst) practice, DMUs are categorized into two sets: non-best and best (or non-worst and worst). In Stage II of best (or worst) practice, the best (or worst) DMUs are further scrutinized with the super-(hypo-) VGA-model to assess their performance differences. Each DMU evaluated by the VGA-models in both Stage I and Stage II is assigned a \textbf{constant} Returns to Virtual Scale (RVS) value. As shown by the blue dashed arrows in Figure \ref{fig1}, the four VGA-methods are identified as: \textbf{b1c, b2c, w1c, and w2c.} 

For example, the b2c VGA-method employs the obPT and osTSc VGA-models, as described by \citet{Liu2025b}. The DMUs are ranked according to their performance in the best (or worst) practice assessments. The highest-ranked DMU on the best practice list might be promoted, while the DMU at the bottom of the worst practice ranking could be considered for removal. After removing these two from the decision matrix, apply the Perspective-c methods again to choose the subsequent candidates.

In \textbf{Perspective-v}, a particular DMUs among the \textit{n} DMUs  aims to compare itself with others to determine its benchmark DMUs by calculating the necessary adjustments in its inputs and outputs to match the performance of the benchmark DMUs. As depicted in Figure \ref{fig1}, the routes marked by red solid arrows, namely \textbf{b1v, b2v, w1v, and w2v}, are to be followed. For instance, the b1v VGA-method uses the PT and TSc VGA-models, introduced by \citet[Figure 1]{Liu2025a}.

 In the initial stage of best (worst) practice, Decision Making Units (DMUs) are categorized into two groups: non-best and best (non-worst and worst). Scenario I employs two VGA-models in a four-phase process to evaluate each non-best (non-worst) DMUs among the \textit{n} DMUs. Each DMU is tasked with selecting the suitable scalar in the variable Returns-to-Virtual-Scale (vRVS) VGA-models in Phase-4 to establish the feasible adjustment rates for both inputs and outputs.

The group of best (worst) DMUs, identified by the Stage I VGA-model in Scenario I is processed by Scenario II, which involves the two super-(hypo-) VGA-models in a four-phase procedure. Each Decision-Making Unit (DMU) within the group is tasked with selecting the suitable scalar during Phase-4 of the variable RVS (vRVS) VGA-models to identify the practically obtainable adjustment rates for both inputs and outputs. For example, the b1v VGA-method employs the sPT and sTSc VGA-models, as introduced by \citet{Liu2025a}.

Even though the \textit{n} DMUs have been evaluated, they cannot be ranked because their subjective preferences in choosing the scalars influence the determination of adjustment rates.

 \begin{figure}[ht!] 
\centering
 \includegraphics[width=1 \textwidth]{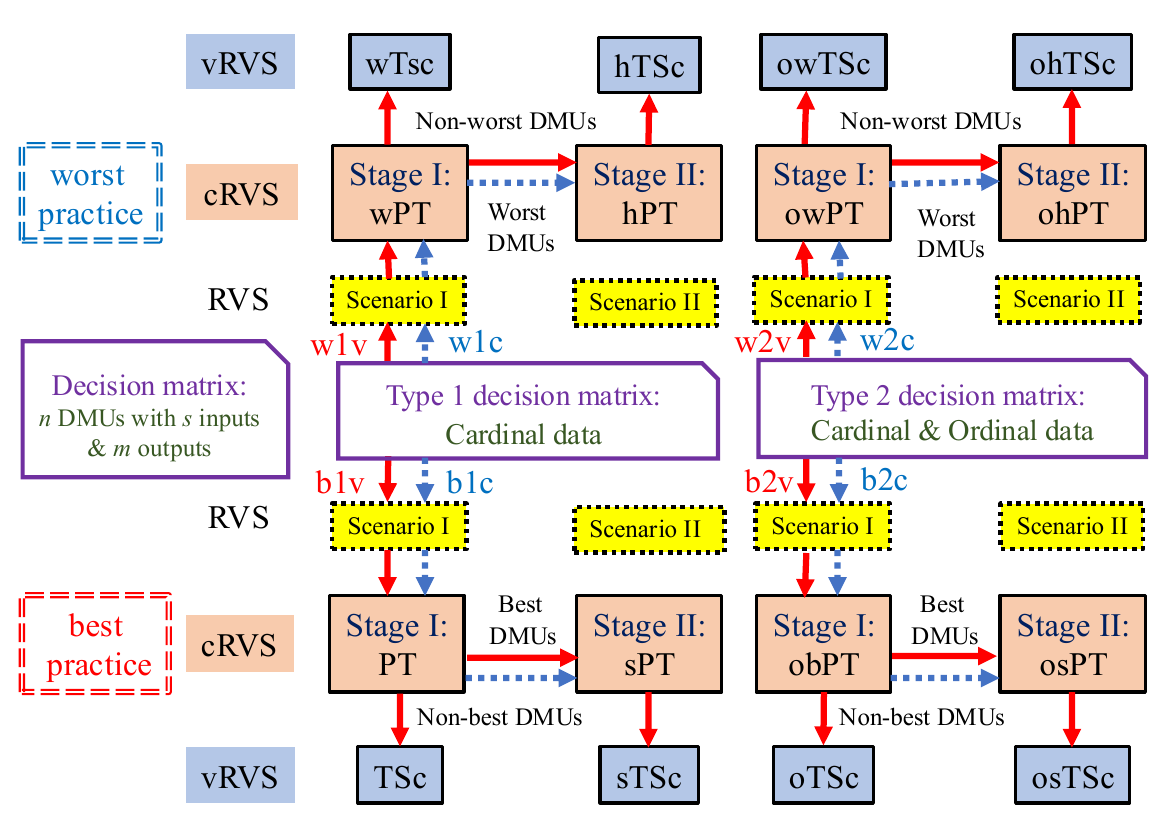}
 \caption {The classifications of VGA-methods and VGA-models.} \label{fig1}
  \end{figure}
\item 
VGA-models:
\begin{itemize}
    \item 
Figure \ref{fig1} illustrates how the b1v VGA-method integrates four distinct VGA-models: Pure Technical (PT), super-Pure Technical (sPT), Technical and Scale choice (TSc), and super-Technical and Scale choice (sTSc), as outlined in \citet{Liu2025a}. This integration is utilized to analyze the Type 1 decision matrix in an optimal practice context.
\item 
The VGA-models associated with the b1v VGA-method, including PT, sPT, TSc, and sTSc, have been revised and now have an 'o' suffix, resulting in models named obPT, osPT, obTSc, and osTSc. These updated models are applied in the b2v VGA-method to evaluate each $DMU_o$ using a Type 2 decision matrix that includes ordinal data.
\item
The VGA-models used in the b1v VGA-method, PT, sPT, TSc, and sTSc have been upgraded to VGA-models, specifically wPT, hPT, wTSc, and hTSc, to form the w1v VGA-method. 
\item 
The VGA-models used in the w1v VGA-method, wPT, hPT, wTSc, and hTSc have been upgraded to VGA-models with an 'o' suffix, specifically owPT, ohPT, owTSc, and ohTSc, to form the w2v VGA-method. 

\item 
The lowercase letters 'b', 's', 'w', and 'h' represent the terms best, super, worst, and hyper, respectively. These VGA-methods are currently under review for publication.
\end{itemize}
\end{enumerate}

\subsection{Ways to eliminate the limitations of existing MCA methods}\label{sec:1.3} 
Over the last fifty years, MCA-related literature, including MCDM, DEA, and SFA methods, boasts over a hundred textbooks and tens of thousands of journal articles. This underscores the importance of both academic research and practical applications. However, the limitations of current MCA methods reduce their applicability.

\citet{Liu2025a,Liu2025b} developed innovative linear programming-based VGA-models designed to address the limitations previously identified in the existing MCA methods. As depicted in Figure \ref{fig1}, the framework consists of 16 VGA-models that proficiently resolve the challenges associated with MCA. 
\begin{enumerate}
\item  
The decision matrix's inherent heterogeneity regarding input and output data does not affect the evaluations.
\item 
Accurately defining parameters associated with criteria ensures that they are not influenced by subjective views, thus fostering consistency.
\item 
The well-defined VGA-methods and VGA-models form a widely recognized framework within MCA.
\item 
Even as decision matrices grow, for instance, reaching 20 or more rows and exceeding 200 columns, it is still feasible to maintain real-time processing.
\item 
The VGA-models and VGA-methods can seamlessly integrate with Artificial Intelligence (AI), the Internet of Things (IoT), and big data.
\item
The Perspective-v VGA-methods engage stakeholders by integrating fairness and bias reduction to achieve practically attainable performance. 
\item VGA-methods can be adapted for real-time settings, such as smart cities or Industry 4.0, where prompt decision-making is essential under uncertain conditions. 
\item The frameworks of the VGA-method can be extended to address real-world challenges involving data that is incomplete, ambiguous, or imprecise.
\end{enumerate}

\subsection{The minimal numerical example}\label{sec:1.4}
Table \ref{table1} is used to illustrate the VGA-method. Table \ref{table1} displays six distinctive laptop models (DMUs) that have quantitative performance indicators \(X_1\) and \(Y_2\), which are characterized as continuous, positive, cardinal data. Within this framework, \(X_1\) denotes the weight of the laptop in kilograms per piece, while \(Y_2\) signifies the quantity of pieces sold. Conversely, the appraisal of qualitative metrics \(X_2\) and \(Y_1\). The combination of quantitative (cardinal) and qualitative (ordinal) measures effectively assesses practical applications.

In Table \ref{table1}, \(X_2\) encapsulates brand perception, with an evaluative point spanning from 1 ("very good") to 6 ("very bad"). In a similar vein, \(Y_1\) pertains to user satisfaction, categorized into four distinct levels ranging from 4 ("very satisfied") to 1 ("very unsatisfied"). The boundaries of the Likert scale for \(X_2\) and \(Y_1\) are specified by \((1_L^x, 6_U^x)\) and \((1_L^y, 4_U^y)\), respectively.

As illustrated in Figure \ref{fig1}, this study introduces the w2c VGA-method. The decision-making team engaged in the Perspective-c evaluation process aims to rank DMUs through a two-stage procedure, employing the \textbf{\textit{ordinal-worst-Pure Technical}} \textbf{(owPT)} and \textbf{\textit{ordinal-hypo-Pure Technical}} \textit{(ohPT)} models.

Table \ref{table1} displays the notations for decision variables utilized in the owPT and ohPT VGA linear programming models. The right-hand column is associated with the \textbf{adjustment rates} and \textbf{virtual unit prices} of the criteria. The bottom row indicates the \textbf{intensities} of DMUs and the objective variables used in both the primal and dual programs. These notations are denoted with the subscript 'o', representing the solutions, which are referred to as \textbf{parameters}, in evaluating each $DMU_o$. The owPT and ohPT VGA-models are the Perspective-c MCA methods that utilize non-parametric approaches in which all parameters are systematically determined.

\begin{table}[!ht]     
     \caption{A basic functional example of a Type 2 decision matrix. Source: \citet{Liu2025b}} 
         \label{table1}
    \centering
     \setlength{\tabcolsep}{2 pt}
\renewcommand{\arraystretch}{1.2}
      \begin{tabular}{llccccccccccll} 
    
      \multicolumn{8}{c}{\textbf{ A minimal illustrative MCA example}} &\multicolumn{4}{c}{\quad} & \multicolumn{2}{c}{\textbf{Decision variables} }\\ \cline{1-8}
\multicolumn{2}{c}{Metrics $\dagger$} &\multicolumn{6}{c}{Decision Matrix} &\multicolumn{4}{c}{\quad} & \multicolumn{2}{c}{of the metrics}\\  
\cline{1-2} \cline{13-14}\cline{9-10}
\multicolumn{2}{c}{\textit{m Inputs } \&}&\multicolumn{6}{c}{{\textit{n} DMUs }} & \multicolumn{4}{l}{\quad}  &Virtual price& Rate of \\ \cline{3-8}
\multicolumn{2}{c}{\textit{s Outputs }}  & K	&	A	&	B	&	D &	G &	H	 &\multicolumn{4}{c}{\quad} & per unit & adjust.\\ \cline{1-8} \cline{13-14}
$X_1$ Input:&$x_{1j}(kg)$   &1.6   &2.3   &1   &1.9   &1.8   &2.5   &\multicolumn{4}{c}{\quad}&$v_{1o}(\$/kg)$&$q_{1o}$ \\
$X_2$ Input:&$x_{2j}(pt.)$   &4   &3   &$6^{xU}$   &5  &3   &$1^{xL}$  &\multicolumn{4}{c}{\quad}&$(v_{2o}\pm d_{2o}^x)(\$/pt.)$&$q_{2o}$ \\
$Y_1$ Output: & $y_{1j}(lvl.)$ &  2   &3   & $1^{yL}$   &1   &2   & $4^{yU}$   &\multicolumn{4}{c}{\quad} & $(u_{1o}\pm d_{1o}^y)(\$/lvl.)$ & $p_{1o}$  \\
 $Y_2$ Output:&$y_{2j}(piece)$   &49   &97   &89   &97   &57   &70  &\multicolumn{4}{c}{\quad}&$u_{2o}(\$/piece$)&$p_{2o}$ \\
\cline{1-8} \cline{13-14}
\multicolumn{2}{c}{\textbf{Decision variables}}&$\pi_{oK}$&$\pi_{oA}$&$\pi_{oB}$ &
$\pi_{oD}$
&$\pi_{oG}$
&$\pi_{oH}$
&\multicolumn{4}{c}{\quad}& \textbf{Decision variables} & $\Delta_o^\star, \delta_o^\star$,\\
\cline{3-8} 
\multicolumn{2}{c}{of the DMUs} & \multicolumn{6}{c}{Intensity of each $DMU_j$ } & \multicolumn{4}{c} {\quad} & of objective values &   $ \tau_o^\star \ddagger$  \\
\cline{1-8} \cline{13-14}
  \multicolumn{14}{l} {$\dagger$ Performance metrics of \textit{m} minimization criteria (inputs)}\\
  \multicolumn{14}{l} {$\qquad$ and \textit{s} maximization criteria (outputs).}  \\
  \multicolumn{14}{l}{$\ddagger$: $\Delta_o^\star$, the total virtual gap price; $ \delta_o^\star$, the total virtual adjustment price; $ \Delta_o^\star$=$\delta_o^\star$.}\\
   \multicolumn{14}{l}{ $\tau_o^\star$, an unified goal price of all metrics; $\$$, a virtual currency.}\\
 \end{tabular}
\end{table} 

\subsection{Organization of the paper}\label{sec:1.5}
The paper is organized as follows: 
\begin{itemize} 
\item Section \ref{sec:2} presents the two-stage framework of the Ord-VGA for ranking the DMUs in worst practice.

\item Section \ref{sec:3} reviews the existing literature on VGA, DEA, SFA, and MCDM methods.

\item Section \ref{sec:4} examines the Ord-VGA-model of Stage I, addressing both primal and dual program duality features.
\item Section \ref{sec:5} describes the Ord-VGA-model of Stage II. 
\item Section \ref{sec:6} explores the derived optimal solutions. 
\item Section \ref{sec:7} provides a simple example and a comprehensive real-world problem \cite{Chen} to demonstrate the advantages of the Ord-VGA. 
\item Section \ref{sec:8} offers conclusions and highlights significant findings. 
\item The appendix includes the dataset for the substantial real-world MCA problem discussed in \citealt{Chen}. 
\end{itemize}

\section{Structure of the novel w2c VGA-method to MCA} \label{sec:2}
 
\begin{figure}[ht!] 
\centering
 \includegraphics[width=0.6 \textwidth]{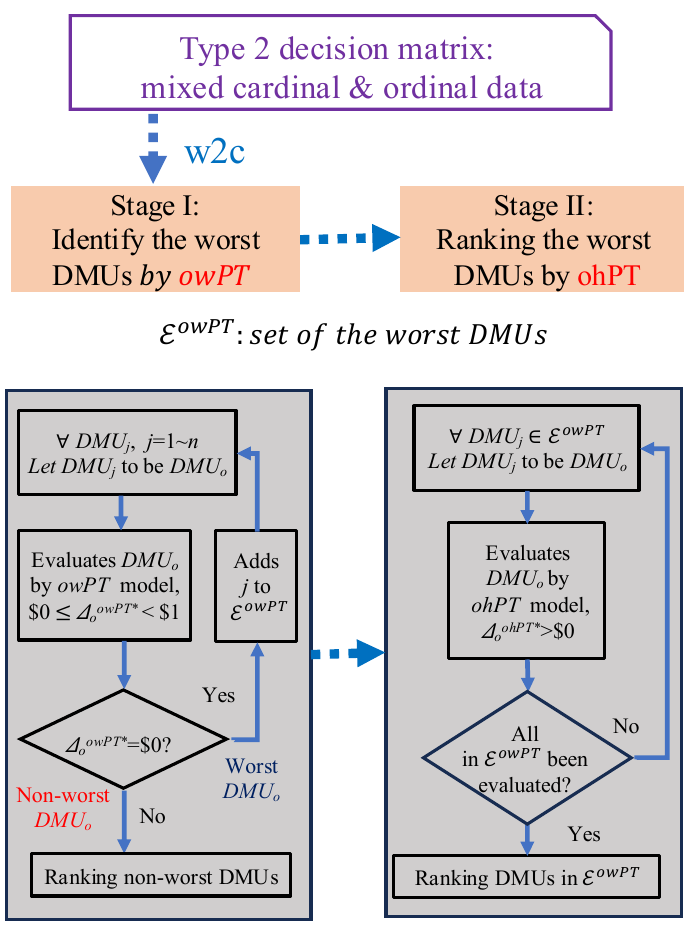} 
 \caption {The w2c VGA-method with the owPT \& ohPT VGA-models.} \label{fig2}
  \end{figure}

As shown in Figure \ref{fig2}, the w2c VGA-method is divided into two stages, which utilize the owPT and ohPT VGA-models. 
\subsection{Symbols and Proper Nouns} \label{sec:2.1} 
Consider the set $J$, comprising $n$ Decision Making Units (DMUs), which are evaluated using quantitative and qualitative performance metrics represented in cardinal and ordinal data sets, denoted as ($I^C, R^C$) and ($I^O, R^O$). The input performance metrics $I^C$ and $I^O$ are classified under the \textit{input metrics}, while $R^C$ and $R^O$ are designated as the \textit{output metrics}.

The decision matrix, $(X, Y)$ consists of row vectors representing inputs and outputs, designated as $X_i$ where $i=1,...,m$ and $Y_r$ where $r=1,...,s$. For any column related to $DMU_j$, the measured values of $X_i$ and $Y_r$ are articulated as $(x_j, y_j)^t$, with $x_{ij}$ and $y_{rj}$ corresponding to particular measurements. The minimum and maximum rankings on the Likert scale for $X_i$ and $Y_r$ are indicated by $(M_i^{xL}, M_i^{xU})$ and $(M_r^{yL}, M_r^{yU})$, respectively.

Every VGA-model consists of two interrelated programs: the primal and the dual. These programs systematically assess each DMUs, denoted as $DMU_o$, relative to all other DMUs by using decision variables with the subscript 'o.' The VGA-model utilizes the following symbols to denote decision variables that are \textit{non-negative}. The numerical example provided in Section \ref{sec:7.1} demonstrates notations consistent with the decision matrix.
\begin{itemize}
    \item ($q_{io}$, $p_{ro}$) – \textbf{Adjustment ratios} (dimensionless) for ($X_i$, $Y_r$), presented as vectors $(Q_o, P_o)$.
    \item ($v_{io}$, $u_{ro}$) – \textbf{Virtual price per unit} (\$) for ($X_i$, $Y_r$), represented in vector form as $(V_o, U_o)$.
    \item ${\pi_{oj}}$ – \textbf{Intensity variable} (dimensionless) for $DMU_j$ assessed in relation to $DMU_o$, shown in vector form as ${\Pi_o}$.
    \item ($d_{io}^x, d_{ro}^y$) – \textit{Virtual unit prices} $(\$)$ to modify ($v_{io}$, $u_{ro}$) for Likert scale values associated with ($X_i$, $Y_r$), with vector expressions ($D_o^x, D_o^y$).
    \item 
    $(x_{io} v_{io}, y_{ro}u_{ro})$ – virtual prices (\$) of (input-i, output-r). 
    \item 
     $(\alpha_o, \beta_o)$: $DMU_o$'s (\textit{virtual input, virtual output})=($\sum_{i\in I} x_{io} v_{io}, \sum_{i\in R} y_{ro}u_{ro})$ (\$).
    \item 
    ($\Delta_o^{owPT}, \Delta_o^{ohPT}$): the virtual gaps of $DMU_o$ in the owPT and ohPT models. 
\end{itemize}

A decision variable denoted by the superscript '*' indicates the value that has been calculated.

\subsection{Stage I process}\label{sec:2.2}
In Stage I, apply the owPT model to assess each $DMU_o$ within the decision matrix, resulting in \textit{n} evaluations needing to be completed. A non-worst $DMU_o$ with a positive virtual gap, $\Delta_o^{owPT\star}>\$0$. When $DMU_o$ simultaneously increases inputs and decreases outputs following the estimated adjustment rates, the virtual gap would diminish to zero. This adjustment would classify it among the worst performers. Out of the non-worst DMUs, the one possessing the largest virtual gap is considered the highest in rank.

The optimization process defines the intensity levels for each $DMU_j$, denoted as $\pi_{oj}^{owPT\star}$. The \textbf{reference peers} of $DMU_o$ are contained within the sets $\mathcal{E}_o^{owPT}$. These peers have $\pi_{oj}^{owPT\star}>0$ and $\Delta_{oj}^{owPT\star}=\$0$, making them the worst DMUs compared to $DMU_o$. In contrast, the \textbf{non-peers} of $DMU_o$ show zero intensity, $\pi_{oj}^{owPT\star}=0$, and a positive value of $\Delta_{oj}^{owPT\star}>\$0$; thus, they are not part of $\mathcal{E}_o^{owPT}$. When $DMU_o$ has $\pi_{oo}^{owPT\star}=1$ and a virtual gap of $\Delta_{oo}^{owPT\star}=\$0$, it is considered the least effective among the n DMUs.

At the end of the first stage, DMUs that exhibit zero virtual gaps, $\Delta_o^{owPT\star}=\$0$, are identified as the worst-DMUs and are grouped into the set $\mathcal{E}^{owPT}$. 

\subsection{Stage II process}\label{sec:2.3}
In Stage II, each DMUs within $\mathcal{E}^{owPT}$ is assessed using the ohPT model in comparison to the other DMUs in $\mathcal{E}^{owPT}$, excluding itself from the comparison. In the ohPT evaluation, the DMU in question, $DMU_o$, has a virtual gap $\Delta_o^{ohPT\star}$ that falls within the interval [\$0, \$1).

In Stage II, DMUs with zero virtual gaps are considered the best among the worst-DMUs. Conversely, a DMU with a positive virtual gap, $\Delta_o^{ohPT\star}>\$0$, termed the \textbf{hypo-virtual gap}, signifies the requirement to simultaneously decrease inputs and increase outputs according to the estimated adjustment rates. 

The optimization procedure determines the intensity levels for each $DMU_j$, represented by $\pi_{oj}^{ohPT\star}$. The \textbf{reference peers} of $DMU_o$ are included in the sets $\mathcal{E}_o^{ohPT}$. These peers have a condition of $\pi_{oj}^{ohPT\star}>0$ and $\Delta_{oj}^{ohPT\star}=\$0$, indicating they are the most efficient DMUs compared to $DMU_o$. Conversely, the \textbf{non-peers} of $DMU_o$ exhibit zero intensity, $\pi_{oj}^{ohPT\star}=0$, and a positive value of $\Delta_{oj}^{ohPT\star}>\$0$, and thus, they are excluded from $\mathcal{E}_o^{ohPT}$. When $DMU_o$ is characterized by $\pi_{oo}^{ohPT\star}=1$ and a virtual gap of $\Delta_{oo}^{ohPT\star}=\$0$, it is regarded as the most efficient entity among the DMUs in $\mathcal{E}_o^{owPT}$.

$DMU_o$ is worse than its reference peers. This process elevates the virtual gap $\Delta_o^{ohPT\star}$ to zero, transforming it into one of the best among the worst-DMUs. The DMU with the greatest virtual gap is ranked at the bottom of the list.

\subsection{Evaluations remain unaffected by decision matrix heterogeneity }\label{sec:2.4}

 Both the inputs and outputs of $DMU_o$ and its reference peers are configured similarly to achieve the optimal virtual gap in the owPT and ohPT models. On the contrary, the non-peer entities of $DMU_o$ exhibit diverse inputs and outputs compared to $DMU_o$. The heterogeneity present within a decision matrix is not easily defined or regulated.

Despite the intrinsic diversity within a decision matrix, the evaluations of DMUs remain unchanged. This represents a fundamental innovation for MCA.   

The two-stage process yields \textit{n} sets of optimal parameters to rank the \textit{n} DMUs. This is the only purely \textbf{non-parametric} approach to solving the MCA issue. A subroutine for the method's flowchart, shown in Figure \ref{fig2}, is created to implement the method, and the use of linear programming software packages allows for quick and complete solutions. 

\subsection{Evaluations are unbiased}\label{sec:2.5} 
Each \textit{variable virtual price} in the owPT and ohPT models is constrained by a specified \textbf{unified goal price} $\tau_o$. A \textbf{two-step approach} was developed to tackle each VGA-model. Step I involves setting the \textit{unified goal price} $\tau_o^\#$ at $\$1$, denoted in virtual currency (\$). Subsequently, the second unified goal price $\tau_o^\star$ for $DMU_o$ is methodically derived from the outcomes of Step I. The decision variables $\tau_o$ are represented with superscripts '$\#$' for the optimal solutions obtained in Step I and '$\star$' for those derived in Step II. After evaluating each $DMU_o$ using the model, the sets of normalized solutions can be compared to one another.

 These virtual input and output pairings for the DMUs are depicted on an innovative two-dimensional intuition graph. This 2D graph shows that the virtual gap for $DMU_o$ is within the range of (\$0, \$1), indicating that its inefficiency score lies between (0,1). Additionally, the graph facilitates the visualization of the reference peers for $DMU_o$. Moreover, the graph illustrates that with the adjusted input and output values for $DMU_o$, a virtual gap does not consistently manifest.

\subsection{Research Principles and Objectives}\label{sec:2.6}
The MCA method, known for its practical application, is based on these key principles: 
\begin{enumerate}
\item Evaluate DMUs using both quantitative and qualitative metrics, depicted through cardinal and ordinal data.
\item Convert each qualitative metric into a Likert Scale using various approaches.
\item Decision matrices feature positive values with defined measurement units.
\item Recognizes and accommodates the inherent \textbf{heterogeneity} among DMUs.
\item Formulates linear programming-based Ord-VGA-models without additional assumptions.
\item Validates the dual attributes of each Ord-VGA-model.
\item Eliminate subjective biases and personal judgments, avoiding normalization, weighting, or uniform criteria among units.
\item The computational process is both efficient and effective, swiftly pinpointing the poorest DMU from extensive decision matrices. A full ranking of the \textit{n} DMUs can be achieved.

\end{enumerate}
Section \ref{sec:7} provides both a succinct example in operation and a detailed real-world scenario. The thorough solutions presented demonstrate that the aims of this study have been met. No other MCA approach has achieved these results.

In scenarios involving MCA issues, computer-based solutions are crucial for prompt resolution and comprehensive report generation. The Ord-VGA framework strategically utilizes the powerful MCA technique to substantially expand its range of applications. Through the development and implementation of highly effective solutions, it adeptly identifies even the least desirable options with accuracy.

Integrating this strategy smoothly into decision-making processes not only enhances system efficiency but also sets a new standard for operational excellence. Additionally, the thoughtful incorporation of side constraints from advanced DEA, SFA, and MCDM approaches into Ord-VGA-models proves to be highly advantageous, resulting in outstanding performance and encouraging innovation.

\section{Literature Review} \label{sec:3} The MCA methods have three categories of approaches: non-parametric, semi-parametric, and parametric.
\subsection{Non-parametric Approaches}\label{sec:3.1}
Non-parametric approaches address the constraints of parametric and semi-parametric MCA techniques.  VGA methods are distinctive non-parametric approaches to MCA.
\subsubsection{VGA-methods evaluates DMUs with cardinal Data}\label{sec:3.1.1}
Professor Shih offered advice on how to incorporate these models into unpublished theses and provided direct assistance in integrating the VGA-model within research efforts that utilize cardinal datasets \citep{JCLi2022, LinZY2024}. Later, \citet{Liu2025a} proposed the b1v VGA-method.

In their work, \citet{Liu2025a} advocates for employing VGA-models to detect \textit{virtual gaps} and to simultaneously tune the inputs and outputs of $DMU_o$. These models excel in establishing a \textbf{unified goal price} in virtual currency, which serves as the standard for the weighted values of inputs and outputs of $DMU_o$, facilitating more detailed and precise analyzes. The primary aim of the VGA-model is the virtual gap, described as the discrepancy between total weighted inputs and outputs. This virtual gap, confined to the interval [\$0, \$1), acts as a unit-less measure of inefficiency.

\subsubsection{VGA-methods evaluates DMUs with Cardinal and Ordinal Data}\label{sec:3.1.2}
\citet{Liu2025b} developed a novel MCA method, the b2c VGA-method, aimed at evaluating and ranking DMUs using both \textbf{cardinal and ordinal data} in practical contexts. Ordinal data, assessable via subjective methods such as group consensus, surveys, or voting, are often presented on Likert scales, which are vital for expressing the unique qualitative features of Decision-Making Units (DMUs). This \textit{best practice} MCA method merges two linear programming-based Ord-VGA-models to address the limitations observed with the VGA approach invented by \citet{Liu2025a}. Importantly, the ranking in a best practice scenario is irreversible and cannot be altered to fit a worst practice scenario.

\subsection{Parametric Approaches}\label{sec:3.2}
\subsubsection
{Stochastic Frontier Analysis Models}\label{sec:3.2.1} 
Statistical SFA is a model for production frontiers aimed at estimating the technology parameter vector related to inputs by using observed input values and a single output from all DMUs. This vector helps in determining the adjusted output and technical efficiency for each DMU. The practical application of SFA is limited as it addresses each output separately. This method is comprehensively explained in a textbook by \citet{Kumbhakar}, with these limitations clearly described by \citet{Theodoridis}. Furthermore, ordinal metrics do not satisfy the statistical requirements for inclusion in SFA models.
\subsubsection{MCDM Methods}\label{sec:3.2.2} 
 Presently, various \textbf{parametric MCDM techniques} determine the weights for inputs and outputs and utilize them to evaluate the performance of alternatives. An MCA decision matrix evaluated through different MCDM methods may yield varied outcomes; yet, their solutions do not show substantial inaccuracies. Selecting the best method is difficult due to the distinct advantages and disadvantages of each.
\citet{Kaya} performed a bibliometric analysis on multi-criteria decision-making (MCDM) methods and reported that 10,387 studies were published from 1974 to 2024. Engineering had the highest number of publications, with the Analytical Hierarchy Process (AHP) and the Technique for Order Preference by Similarity to Ideal Solution (TOPSIS) being the most commonly applied concepts. Other popular methods include VIKOR, ELECTRE, and FUCOM. He revealed the primary weakness of MCDM methods. 

 \citet{Amor2023} through the bibliometrix tool indicated that contemporary MCDM methods often yield divergent outcomes, prompting scholarly debate. Similarly, \citet{Danielson2022} delved into the inconsistencies associated with MCDM ranking methods. 

\citet{Taherdoost} performed a comprehensive assessment of MCDM, discussing its foundational concepts, applications, principal categories, and methodologies. Certain MCDM techniques endeavor to integrate both cardinal and ordinal data in criterion weight evaluations; yet, they face difficulties with differing measurement units. 

\citet{Antunes} indicated that the influence of assumptions embedded in various MCDM models on the resulting scores is frequently overlooked. To tackle this oversight, a hybrid DEA-MCDM methodology was suggested, which allows for the assessment of both DMU performance and synergy, facilitating their categorization based on synergy performance tiers. 

\subsection{Semi-parametric Approaches}\label{sec:3.3}
\subsubsection{Initial Advances in VGA-models}\label{sec:3.3.1}
\citet{Liu2015} made noteworthy contributions to the development of Virtual Gap Measurement (VGM) models tailored for use with cardinal data, where the \textit{unified virtual goal price} was subjectively established, leading to partial evaluations. Building on this work, as demonstrated by the study conducted by \citet{Liu2017a, Liu2017b}, the VGM model was applied to two-stage production systems with cardinal datasets.
\subsubsection{Data Envelopment Analysis (DEA) methods} \label{sec:3.3.2}
\begin{enumerate}
    \item 
    The development of DEA theory:
    
 Besides the parametric MCDM methods, \citet{Charnes} introduced DEA models grounded in linear programming to tackle MCA utilizing semi-parametric methods. Each DEA model assesses a specific $DMU_o$ by comparing it to other DMUs. The \textit{n} DMUs alternately serve as $DMU_o$. DEA theory posits an efficiency frontier that encompasses DMU points within a hyperspace of \textit{(m+s)} dimensions.

The \textbf{additive DEA model} aims to enhance efficiency by both reducing inputs and increasing outputs to achieve optimal efficiency ratings for \( DMU_o \), applicable under conditions of constant returns-to-scale (CRS) and variable returns-to-scale (VRS). The VRS model has an additional dual decision variable with positive or negative values, $w_o$.
In DEA literature, additive CRS and VRS models hold significant importance. Radial efficiency DEA models, whether focused on inputs or outputs, do not completely assess each $DMU_o$. 
\item
Articles that improve DEA theory frequently appear in academic journals and textbooks to address the limitations of MCDM methods. Nonetheless, DEA models struggle to achieve acceptable evaluations due to substantial theoretical weaknesses.
\item 
The problematic DEA theory:

In each additive model's dual program, the constraints $(v_{io}\leq b_{ix},\forall i\in I, u_{ro}\leq b_{ry}\forall r\in R)$ are included. Here, $(v_{io},\forall i\in I, u_{ro}\forall r\in R)$ represents the variable weights, while $(b_{ix},\forall i\in I, b_{ry}\forall r\in R)$ is termed \textbf{artificial goal weights}, as determined by the model developer. In DEA literature, several developers have endeavored to identify suitable artificial goal weights \citep{Halicka}. DEA is considered a \textbf{semi-parametric} method because it uses a set of \textbf{artificial goal weights} to evaluate each $DMU_o$. 

 The two objective functions obtained the minimized dimensionless inefficiency scores of $DMU_o$.
\begin{align*}
F_o^{CRS\star}&=\sum_{i\in I} x_{io} v_{io}^{\star}- \sum_{i\in R} y_{ro}u_{ro}^{\star}\\
F_o^{VRS\star}&=\sum_{i\in I} x_{io} v_{io}^{\star}- \sum_{r\in R} y_{ro}u_{ro}^{\star}-w_o^\star
\end{align*}
Unlike articles that review and compare MCDM methods, each DEA model exposes its own errors. The significant errors in the DEA formulations are:
\begin{itemize}
\item 
The units of measurement in the objective functions are inconsistent, which goes against the principle of formulating a linear programming model  \citep{Bazaraa}. It is necessary to define and maintain consistency in the measurement units of decision variables, constraints, and data utilized in both primal and dual programs.
\item 
$F_o^{CRS\star}=\sum_{i\in I} x_{io} v_{io}^{\star}- \sum_{r\in R} y_{ro}u_{ro}^{\star}$ should be equal to $F_o^{CRS\star}=1-E_o^{CRS\star}=1-(\sum_{r\in R} y_{ro}u_{ro}^{\star}/\sum_{i\in I} x_{io} v_{io}^{\star})$. The condition holds only if $\sum_{\forall i \in I }x_{io} v_{io}^{\star}=1.$ 
\item
None of the existing artificial goal weights could achieve the condition   $\sum_{\forall i \in I }x_{io} v_{io}^{\star}=1.$
\item 
\Citet{Halicka} studied that there is no feasible way to have a $E_o^{VRS\star}$ or $F_o^{VRS\star}$.
\item 
The 32 footnotes in \citet{Liu2025a} and the literature review in \citet{Liu2025b} highlight additional errors inherent in DEA theory.
\end{itemize}
\item 
DEA models for worst practice and hypo-efficiency:

In the conventional DEA context, the worst practice DEA model, referred to as the \textit{inverse DEA model}, operates by increasing inputs and decreasing outputs to determine the efficiency score in relation to a worst practice frontier (WPF) \cite{Liu2009, Liu2012}. Unlike the optimistic best practice models, this inverse method is viewed as pessimistic.

An adjusted DEA model assigns a \textit{super-efficiency} score that can exceed 1, compared to the traditional DEA model.  While the principle of super-efficiency utilized in best practices can theoretically be applied to worst practices as hypo-efficiency, there is currently no documented evidence of hypo-efficiency models in the available literature.

\item 
Issues of Heterogeneity in DMUs:

All DEA models presuppose that DMUs are uniform. As highlighted by \cite{Aleskerov, Zarrin, Zhu}, there are concerns regarding the composition of the DMUs' within the decision matrix. The complexity of this issue is highlighted by an extensive body of research over the past years.  However, the set of artificial goal weights in each DEA model falls short in guaranteeing that the inefficiency score of $DMU_o$ remains within the feasible range of [0,1). 

\item 
DEA Models for handling ordinal Data:

\citet{Ebrahimi2024} undertook a comprehensive analysis of DEA models that deal with ordinal data. Nonetheless, numerous ordinal DEA models referenced in \cite{ Chen, Ebrahimi2024} fail to guaranty precise efficiency score estimation. 

\citet{Chen} carried out an assessment employing three diverse DEA models to ascertain the output adjustment ratios. The results of this evaluation are compiled in three distinct tables. Each output was uniformly weighted at one-third. The discussion found in Section \ref{sec:7.2} explores the problem and provides solutions.
\end{enumerate}

\section{Stage I: Identifying the worst- and non-worst DMUs} \label{sec:4}
Stage I evaluates DMUs and partitions them into the worst and non-worst DMUs.

\subsection{The owPT Model.} \label{sec:4.1} 
 The \textbf{dual program} pertinent to the owPT model computes the maximum \textit{total adjustment price} (TAP) $\delta_o^{owPT\star}$ for any specified $DMU_o$ under the worst practice. Equations (\ref{Eq:2}) and (\ref{Eq:3}) quantify the expansion and contraction for each input and output value, respectively. In other words, the inputs and outputs are maximization and minimization criteria, respectively. In parallel, Equations (\ref{Eq:4}) and (\ref{Eq:5}) impose constraints on the adjusted inputs and outputs within their respective positions on the Likert scale. Equations (\ref{Eq:2}) to (\ref{Eq:5}) feature formulas preceded by the symbol $\Leftrightarrow$ in brackets, indicating the original conditions.

 \textbf{TAP model}
\begin{equation} \begin{aligned} \label{Eq:1}	
    \delta_o^{owPT\star} (\$)= \max_{Q_o , P_o , \Pi_o } \sum_{\forall i\in I^C \cup I^O} q_{io} \tau_o  &+ \sum_{\forall r\in R^C \cup R^O}p_{ro} \tau_o \quad \forall o \in J
    \end{aligned}
 \end{equation}
 \quad \quad subject to:
\begin{equation} 
\begin{aligned}
 v_{io}: &-\sum_{\forall j \in J} x_{ij} \pi_{oj} + q_{ro} x_{io} = - x_{io} \quad \forall i \in I^C \cup I^O\\
 &\Leftrightarrow [\sum_{\forall j \in J} x_{ij} \pi_{oj} = (1+q_{ro}) x_{io}]\quad \forall i \in I^C \cup I^O	
\label{Eq:2}
\end{aligned}
 \end{equation}
\begin{equation} \begin{aligned}
u_{ro}: & \sum_{\forall j \in J} y_{rj} \pi_{oj} + p_{ro} y_{ro} =  y_{ro} \quad  \forall r \in R^C \cup R^O \\
&\Leftrightarrow  [\sum_{\forall j \in J} y_{rj} \pi_{oj} =(1- p_{ro}) y_{ro}] \quad  \forall r \in R^C \cup R^O	
\label{Eq:3}
\end{aligned}
 \end{equation}
 \begin{equation} \begin{aligned}
d_{io}^x:&\quad q_{io} x_{io} \leq (M_i^{xU}-x_{io}) \quad \forall i \in I^O	\\
&\Leftrightarrow  \quad [(1+q_{io}) x_{io}  \leq M_i^{xU}]  \quad \forall i \in I^O
\label{Eq:4} \end{aligned}
 \end{equation}
 \begin{equation}\begin{aligned}
d_{ro}^y:& \quad  p_{ro} y_{ro}\leq -(M_r^{yL}- y_{ro})\quad \forall r \in R^O	\\
&\Leftrightarrow  \quad [(1-p_{ro}) y_{ro}\geq M_r^{yL}] \quad \forall r \in R^O
\label{Eq:5}
 \end{aligned}
 \end{equation}
\begin{equation} \label{Eq:6}
\Pi_o , Q_o, P_o  \ge 0	
 \end{equation}

Following the principles of linear programming, the dual program is converted into the \textbf{primal program}, which seeks to determine the minimum \textit{total virtual gap} (TVG) $\Delta_o^{owPT\star}$ for $DMU_o$.

\textbf{TVG model}
 \begin{equation}
\begin{aligned}
	\Delta_o^{owPT\star} (\$)&= \min_{V_o, U_o, D_o^x, D_o^y} 
    -\lbrace\sum_{\forall i\in I^C} v_{io} x_{io}+\sum_{\forall i\in I^O}[v_{io}x_{io}+(M_i^{xU}-x_{io})d^x_{io}]\rbrace\\
    &+ \lbrace \sum_{\forall r\in R^C } u_{ro} y_{ro} + 
     \sum_{\forall r\in R^O} [u_{ro}y_{ro}-(M_r^{yL}-y_{ro})d^y_{ro}]\rbrace  
     \quad \forall o \in J	
 \label{Eq:7}
 \end{aligned}
 \end{equation}
 \quad \quad subject to:
 \begin{equation}\begin{aligned}	
    \quad \pi_{oj}: -\sum_{\forall i\in I^C\cup I^O} v_{io} x_{ij}  + \sum_{\forall r\in R^C \cup R^O}u_{ro}  y_{rj}  \geq 0(\$)\quad \forall j\in J \label{Eq:8}
    \end{aligned}
 \end{equation}
 \begin{equation}
	q_{io}: x_{io} v_{io} \geq \tau_o(\$)\quad \forall i\in I^C	
 \label{Eq:9}
 \end{equation}
 \begin{equation}
	q_{io}: x_{io} (v_{io}+d_{io}^x) \geq \tau_o(\$)\quad \forall i\in I^O	
 \label{Eq:10}
 \end{equation}
\begin{equation}\label{Eq:11}
	p_{ro}: y_{ro} u_{ro}\geq \tau_o(\$)\quad \forall r \in R^C
 \end{equation}\begin{equation}\label{Eq:12}
	p_{ro}: y_{ro} (u_{ro}+ d_{ro}^y) \geq \tau_o(\$)\quad \forall r \in R^O
 \end{equation}
\begin{equation}
	V_o, U_o \quad free,\quad  D_o^x, D_o^y \ge 0	
 \label{Eq:13}
 \end{equation}

 Each objective function is delineated over a specific domain, $o \in J$, ensuring that each constraint initiates with its respective decision variable, thereby preserving consistency throughout program transformations.

The \textbf{owPT} model assesses the pure technical virtual gap for $DMU_o$, expressed as ($\Delta_o^{owPT\star}\geq \$0$). The element $DMU_j$ is included in the set of \textit{reference peers} of $DMU_o$, represented by the symbol $\mathcal{E}_o^{owPT}$, if the optimal solution of Equation (\ref{Eq:8}) is precisely zero. While the corresponding dual variable $\pi_{oj}^\star$ is a positive value. The values ($V_o^\star, U_o^\star$) represent the estimated unit prices of inputs and outputs used by reference peers in their technical processes to convert maximization inputs into minimization outputs. 
 
Equations (\ref{Eq:9})-(\ref{Eq:12}) establish a lower boundary for \textit{virtual prices}, referred to as the \textit{unified goal price} $\tau_o (\$)$. During Steps I and II, $\tau_o$ in this context is substituted by $\tau_o^\# = \$1$ and $\tau_o^\star = \$ \bar{t}$, respectively.
The solutions associated with Equation (\ref{Eq:1}) for both Step I and Step II are thoroughly presented in Equation (\ref{Eq:14}), while Equation (\ref{Eq:15}) delineates the solutions pertaining to Equation (\ref{Eq:7}), relevant to both aforementioned steps. In Equation (\ref{Eq:15}), the prices for virtual gaps are shown as pairs of virtual input (\textit{vInput}) and virtual output (\textit{vOutput}), represented by $(\alpha_o^\#, \beta_o^\#)$ and $(\alpha_o^\star, \beta_o^\star)$. In Step I, the overall virtual gap $\Delta_o^{owPT\#}$ can surpass \$1. This total virtual gap comprises four unique elements.
 \begin{equation}
 \begin{aligned}
\$0 \leq &\delta_o^{owPT\#} (\$)= \delta_{xo}^\# +\delta_{yo}^\# = \sum_{\forall i\in I^C \cup I^O} q_{io} ^\# \times \$1  + \sum_{\forall r\in R^C \cup R^O}p_{ro} ^\#  \times \$1\\ 
\$0 \leq & \delta_o^{owPT\star } (\$)= \delta_{xo}^\star +\delta_{yo}^\star =\sum_{\forall i\in I^C \cup I^O} q_{io}^\star  \tau_o ^\star  + \sum_{\forall r\in R^C \cup R^O} p_{ro}^\star  \tau_o ^\star \leq \$1
  \end{aligned}  \label{Eq:14} 
 \end{equation}  
 \begin{equation} \label{Eq:15}
\begin{aligned}
\$0 &\leq  \Delta_o^{owPT\#}(\$) =-(vInput^\#)+(vOutput^\#) = -\alpha_o^\# + \beta_o^\#\\
&=-[V_o^\# x_o +\sum_{\forall i\in I^O} (M_i^{xU}-x_{io}) d_{io}^{x\#}] +[U_o^\# y_o -\sum_{\forall r\in R^O} (M_r^{yL}-y_{ro}) d_{ro}^{y\#}]
\\
\$0 &\leq \Delta_o^{owPT\star }(\$)=-(vInput^\star)+(vOutput^\star)=-\alpha_o^\star +\beta_o^\star\\
&=-[V_o^\star x_o +\sum_{\forall i\in I^O} (M_i^{xU}-x_{io}) d_{io}^{x\star}] +[U_o^\star y_o -\sum_{\forall r\in R^O} (M_r^{yL}-y_{ro}) d_{ro}^{y\star}]\leq \$1
\end{aligned}
 \end{equation}
Equation (\ref{Eq:16}) demonstrates how the solutions derived from Equation (\ref{Eq:8}) for each $DMU_j$ during Steps I and II are merged into the pairs (\textit{vInput, vOutput}), namely $(\alpha_{oj}^\#, \beta_{oj}^\#)$ and $(\alpha_{oj}^\star, \beta_{oj}^\star)$. It is noteworthy that in Step I, $\Delta_{oj}^{owPT\#}$ can exceed \$1.
\begin{equation} \label{Eq:16}
\begin{aligned}
\$0 \leq &\quad  \Delta_{oj}^{owPT\#} (\$)= -V_o^\# x_j +U_o^\# y_j = -\alpha_{oj}^\# + \beta_{oj}^\# \quad \forall j \in J\neq o\\
\$0 \leq &\quad \Delta_{oj}^{owPT\star} (\$)= -V_o^\star x_j +U_o^\star y_j = -\alpha_{oj}^\star +\beta_{oj}^\star\leq \$1 \quad \forall j \in J\neq o 
\end{aligned}
 \end{equation}
 
\subsection{Establish a Unified Goal Price for \texorpdfstring{$DMU_o$}.}  \label{sec:4.3}
In Step II, it is crucial for $DMU_o$ to verify that $\delta_o^{owPT\star}$ and $\Delta_o^{owPT\star}$ reside within the interval of [$\$0, \$1$). The relationships between the owPT model, as described in Steps I and II, are detailed below.
\begin{equation}\begin{aligned} \label{Eq:17}
\tau_o^\# :\tau_o^\star   &=
 1 :  \bar{t} =  \delta_o^{owPT\#}:\delta_o^{owPT\star} =\Delta_o^{owPT\#}:\Delta_o^{owPT\star}\\
 &= (-\alpha_o^\#+ \beta_o^\# ) : (-\alpha_o^\star+\beta_o^\star)
 \end{aligned} \end{equation}
Therefore, the following \textit{relational equation} is valid.
\begin{equation} \label{Eq:18}
\bar{t} \$ (-\alpha_o^\#+ \beta_o^\# ) = 1\$(-\alpha_o^\star+\beta_o^\star) 
\end{equation}
Using Equation (\ref{Eq:19}), we obtained the dimensionless parameter $\bar{t}$.
\begin{equation} \label{Eq:19}
	\$\bar{t} = \$1/\beta_o^\# \textrm {  and  } \tau_o^\star   = \$ \bar{t} 	
  \end{equation}
Equation (\ref{Eq:18}) may be reformulated as Equation (\ref{Eq:20}).
\begin{equation} \begin{aligned} \label{Eq:20} 
 (\$1/\beta_o^\#) \$(-\alpha_o^\#+ \beta_o^\# ) = 1\$(-\alpha_o^\star+\beta_o^\star) \Leftrightarrow 
  \$(   1- \beta_o^\#/\alpha_o^\# ) = \$(-\alpha_o^\star+\beta_o^\star)
 \end{aligned} \end{equation}
 By dividing Equation (\ref{Eq:20}) by $\beta_o^\star$, we obtain:
 \begin{equation}\label{Eq:21}
  (\$1/\beta_o^\star)( 1- \beta_o^\#/\alpha_o^\# ) =(-\alpha_o^\star/\beta_o^\star+1 )
  \end{equation}
  Equation (\ref{Eq:15}) demonstrated that $ \$0 \leq( -\alpha_o^\#+ \beta_o^\# )$ and $ \$0\leq (-\alpha_o^\star+\beta_o^\star) $, which correspond to the following equations:
  \begin{equation} \label{Eq:22}
  0\leq( 1- \alpha_o^\#/\beta_o^\# )\leq 1 \text{ and } 0\leq (1-\alpha_o^\star/\beta_o^\star )\leq 1
  \end{equation}
The parameter $\beta_o^\star = \$1$ identified in Equation (\ref{Eq:21}) confirms the formulas presented in Equation (\ref{Eq:22}), with $\Delta_o^{owPT\star}$ anticipated to fall within the range of [$\$0, \$1$). In Equation (\ref{Eq:23}), it is demonstrated that during Steps I and II, the \textit{virtual inefficiency scores}, denoted by $F_o^{owPT\star}$, are computed without reference to the unified virtual price and are confined to the interval [0, 1). In Step I, each $DMU_o$ impartially employs $\tau_o^\#=\$1$. These steps aim to identify $\tau_o^\star$ through an impartial evaluation process.
  \begin{equation}
\begin{aligned}\label{Eq:23} 
	0\leq F_o^{owPT\star}=\Delta_o^{owPT\star}/\beta_o^\star =  \alpha_o^\#/\beta_o^\#= \alpha_o^\star/\beta_o^\star < 1 
\end{aligned}
\end{equation}

In most scholarly discussions, efficiency analysis often hinges on directly computing the efficiency score. To clarify, the owPT model evaluates the performance of $DMU_o$ through a two-step process, as described in Equation (\ref{Eq:23}), guaranteeing that $\$0 \leq \Delta_o^{owPT\star} < \$1.$

In mathematical optimization, a common method involves normalizing Step I solutions to achieve Step II solutions. Equation (\ref{Eq:24}) presents the virtual gap for $DMU_o$, denoted as $\Delta_o^{owPT\star}$ in Step II, which is standardized and bound within the range [\$0, \$1). This process might include scaling or altering Step I solutions to meet the conditions necessary for Step II. This normalization allows for straightforward comparisons across the \textit{n} evaluations of distinct $DMU_o$.
\begin{equation} \begin{aligned}
 (Q_o^\star, P_o^\star,  \Pi_o^\star) &= ( Q_o^\#, P_o^\#, \Pi_o^\star)\\
  ( \Delta_o^{owPT\star },  V_o^\star, U_o^\star, D_o^{x\star}, D_o^{y\star})
 & =\bar{t} (\Delta_o^{owPT\#}, V_o^\#, U_o^\#,  D_o^{x\#}, D_o^{y\#})
\label{Eq:24} \end{aligned}
\end{equation}

When evaluating $DMU_o$ in Steps 1 and 2, the set of reference peers is $\mathcal{E}_o^{owPT}$. The collective group of the worst DMUs comprises the union of the \textit{n} reference peer sets, as represented by the following equation:
\begin{equation} \label{Eq:25}
    \mathcal{E}^{owPT}= \cup_{\forall o \in J} \quad\mathcal{E}_o^{owPT}
\end{equation}

\subsection{Duality Properties}\label{sec:4.4}
We develop VGA-models utilizing the principles of linear programming theory \citep{Bazaraa}. This involves ensuring that the duality relationship between every pair of primal and dual programs is verified. It is crucial that constraints and decision variables maintain consistent definitions and corresponding measurement units in both formulations.

When \textit{the total virtual gap} (TVG) and \textit{total adjustment price} (TAP) reach their optimal values in the owPT model, as outlined in Equation (\ref{Eq:26}), it indicates that the model's conditions align with its objectives. This parity reveals a harmony between TVG and TAP, highlighting a unified relationship between gap and price metrics.
\begin{equation}\begin{aligned} \label{Eq:26}
\delta_o^{owPT\star}(\$)=\Delta_o^{owPT\star} (\$)
  \end{aligned} \end{equation}

The Ord-VGA-model utilizes linear programming to iteratively pursue the objectives of both the primal and dual programs. It examines the optimized outcomes such as ($V_o^\star, U_o^\star$), ($Q_o^\star, P_o^\star$), ($D_o^{x\star}, D_o^{y\star}$), and $\Pi_o^\star$. Furthermore, verifying the robust complementary slackness conditions for these programs is crucial.
 
\subsection{Strong Complementary Slackness Conditions (SCSC)}\label{sec:4.5}
Equations (\ref{Eq:27}), (\ref{Eq:28}), and (\ref{Eq:29}) illustrate the conditions for the TAP program.
\begin{equation} \begin{aligned} \label{Eq:27}
\relax [\sum_{\forall j\in \mathcal{E}_o^{owPT}} x_{ij} \pi_{oj}^\star - x_{io} (1 + q_{io}^\star)] \times v_{io}^\star =\$0 \quad \forall i\in I^C \cup I^O  
\end{aligned}  \end{equation}
\begin{equation} \begin{aligned}
\relax [\sum_{\forall j\in \mathcal{E}_o^{owPT}} y_{rj}\pi_{oj}^\star - y_{ro}(1-p_{ro}^\star )] \times u_{ro}^\star &=\$0 \quad 
\forall r\in R^C \cup R^O
 \label{Eq:28} \end{aligned} \end{equation}
 \begin{equation} \begin{aligned}\label{Eq:29}
    [(1+q_{io}^\star) x_{io}-M_i^{xU}]  \times d_{io}^{x\star}&=\$0\quad \forall i \in I^O\\ [M_r^{yL}-(1-p_{ro}^\star) y_{ro}] \times d_{ro}^{y\star}&=\$0\quad \forall r \in R^O  
\end{aligned}\end{equation} 
\indent The TVP program satisfies the criteria indicated in Equation (\ref{Eq:30}) and Equation (\ref{Eq:31}).
\begin{equation}
	(-V_o^\star x_j +U_o^\star y_j ) \times \pi_{oj}^\star = \$0\quad \forall j\in J 
	\label{Eq:30}
\end{equation} 
\begin{equation}\begin{aligned}\label{Eq:31} 
	(v_{io}^\star x_{io} - \tau_o^\star)  \times q_{io}^\star &= \$0\quad \forall i \in I^C\\
    \quad [(v_{io}^\star +d_{io}^{x\star}) x_{io} - \tau_o^\star ]  \times q_{io}^\star &= \$0\quad \forall i \in I^O \\
    (u_{ro}^\star y_{ro} - \tau_o^\star) \times  p_{ro}^\star&= \$0\quad \forall r\in R^C\\\quad [(u_{ro}^\star+d_{ro}^{y\star}) y_{ro} - \tau_o^\star ] \times p_{ro}^\star&= \$0\quad \forall r\in R^O 
   \end{aligned} \end{equation}

 In Equations (\ref{Eq:27})-(\ref{Eq:31}), one of the two terms on the left side must become zero in order for their product to equal zero. The Step II calculation process governs this portion. To obtain the conditions for Step I, one must replace the superscript '$\star$' with '$\#$' in each decision variable and substitute $\tau_o^\star$ with $\$1$. 

 In Equation (\ref{Eq:29}), when the ordinal data $x_{io}$ is calibrated between two neighboring Likert points, such that $(1+q_{io}^\star) x_{io}<M_i^{xU}$, we find that $d_{io}^{x\star}=0$, and the virtual unit price becomes $v_{io}^\star$. Conversely, if $x_{io}$ reaches the uppermost point, where $(1+q_{io}^\star) x_{io}=M_i^{xU}$, the virtual unit price changes to $v_{io}^\star+d_{io}^{x\star}>0.$ This leads to a penalty price $(M_i^{xU}-x_{io})d_{io}^{x\star}$ as shown in Equation (\ref{Eq:15}). The explanations related to ordinal data $y_{ro}$ are not reiterated.

\subsection{Sensitivity Analysis} \label{sec:4.6}

In the context of the TVG framework, $V_o$ and $U_o$ denote variables associated with free indices. Equations (\ref{Eq:27}) and (\ref{Eq:28}) demonstrate that their left-hand sides equate to zero, indicating \(V_o^\star >0\) and \(U_o^\star >0\). Given the conditions \( (\textit{X, Y}), (D_o^x, D_o^y)>0\) and \(\tau_o^\star >\$0\), it is confirmed through Equation (\ref{Eq:31}) that the computed virtual unit prices result in \(V_o^\star >0\) and \(U_o^\star >0\). The owPT model accurately estimates $p_{io}^\star$ and $q_{ro}^\star$, ensuring that $DMU_o$ corresponds to the \textit{worst-practice prime Meridian} illustrated in Section \ref{sec:7.1.1}, where \( \beta_o^\star \) is set to \(\$1\).

All $DMU_j$ adhere to the conditions outlined in Equation (\ref{Eq:30}). If $(-V_o^\star x_j + U_o^\star y_j) = 0$ and \(\pi_{oj}^\star > 0\), then $DMU_j$ is considered to belong to \(\mathcal{E}_o^{owPT}\). Those figures  illustrate that the \textit{worst-practice prime Meridian} is the diagonal line at the origin, characterized by \((-V_o^\star x_j + U_o^\star y_j) = \$0\) for every $j$ in \(\mathcal{E}_o^{owPT}\). When $(-V_o^\star x_o + U_o^\star y_o) = 0$, $DMU_o$ achieves a zero total virtual gap, denoted as $\Delta_o^{owPT\star} = \$0$, with $\pi_{o}^{\star} = 1$. 

If $DMU_j$ meets the criteria $(-V_o^\star x_j + U_o^\star y_j) > 0$ and \(\pi_{oj}^\star = 0\), it is omitted from \(\mathcal{E}_o^{owPT}\). As demonstrated in the figures of Section \ref{sec:7.1.1}, $DMU_j$ exhibits an overall positive virtual gap indicated by $\Delta_{oj}^{owPT\star} > \$0$, with $\pi_{o}^{\star} = 0$. Thus, $DMU_j$ is situated above the prime Meridian. The removal of $DMU_j$ from the decision matrix does not influence the evaluation of $DMU_o$.

\section{Stage II: Identifying the worst DMU} \label{sec:5} In Stage I, the DMUs in the subordinate tier are evaluated with a zero virtual gap, which belongs to the $\mathcal{E}^{owPT}$ group, as described in Equation (\ref{Eq:25}). In Stage II, apply the ohPT VGA-model to assess the \textbf{hyper-virtual gap} for each DMU within the $\mathcal{E}^{owPT}$ group. We introduce a technique for identifying the worst DMU based on the calculated hyper-virtual gaps.

Equations (\ref{Eq:33}), (\ref{Eq:34}), and (\ref{Eq:39}) exclude reference $DMU_o$ and stipulate that each $DMU_j$ must demonstrate a non-positive virtual gap. Equation (\ref{Eq:43}) is employed to ascertain the largest hyper-virtual gap of $DMU_o$ concerning other peers. Within the dual program, Equations (\ref{Eq:38}) and (\ref{Eq:39}) elucidate the manner in which $DMU_o$ determines both a reduction rate and an expansion rate for each input and output measure via a linear combination of DMUs, excluding $DMU_o$ itself. In this model, the inputs and outputs are minimization and maximization criteria, respectively.

Unlike Section \ref{sec:3.1}, the symbols $Q_o$ and $P_o$ denote the expansion and contraction ratios of the input and output metrics. Equations (\ref{Eq:40}) and (\ref{Eq:41}) limit the decrease in $x_{io}$ to the lower threshold $M_i^{xL}$ and confine the increase in $y_{ro}$ to the upper threshold $M_r^{yU}$. These equations are related to the decision variables $d_{io}^x$ and $d_{ro}^y$ used within the primate framework. In linear programming, the dual formulation's constraints and objective function can be transformed into a primal formulation and vise versa, providing insight into the principles of duality.

 \subsection{The ohPT Model} \label{sec:5.1} 
 \textbf{The dual program aims to have the total adjustment price (TAP):}
\begin{equation} \begin{aligned} \label{Eq:32}	
    \delta_o^{ohPT\star} (\$)&= \min_{Q_o, P_o, \Pi_o} \sum_{\forall i\in I^C \cup I^O} q_{io} \tau_o  + \sum_{\forall r\in R^C \cup R^O }p_{ro} \tau_o \quad \forall o \in \mathcal{E}^{owPT}
    \end{aligned}
 \end{equation}
\quad \quad subject to:
 \begin{equation} \begin{aligned} 
 \quad v_{io}: &\sum_{\forall j \in \mathcal{E}^{owPT} \neq o} x_{ij} \pi_{oj} + q_{ro} x_{io} \geq  x_{io} \quad \forall i \in I^C \cup I^O\\
 &\Leftrightarrow \sum_{\forall j \in \mathcal{E}^{owPT} \neq o} x_{ij} \pi_{oj} \geq (1- q_{ro}) x_{io} \quad \forall i \in I^C \cup I^O	
\label{Eq:33}
\end{aligned}
 \end{equation}
\begin{equation} \begin{aligned} 
u_{ro}:  &-\sum_{\forall j \in \mathcal{E}^{owPT} \neq o} y_{rj} \pi_{oj} + p_{ro} y_{ro} \geq - y_{ro}\quad \forall r \in R^C \cup R^O\\
&\Leftrightarrow \sum_{\forall j \in \mathcal{E}^{owPT} \neq o} y_{rj} \pi_{oj} \leq  (1+ p_{ro}) y_{ro}
\quad \forall r \in R^C \cup R^O	
\label{Eq:34}
\end{aligned}
 \end{equation}
 
 \begin{equation} 
d_{io}^x:  -q_{io} x_{io} \geq (M_i^{xL}-x_{io} )   \quad \Leftrightarrow \quad [(1-q_{io}) x_{io} \geq M_i^{xL}] \quad \forall i \in I^O	
\label{Eq:35}
 \end{equation}
 \begin{equation} 
d_{ro}^y:  -p_{ro} y_{ro}\geq -(M_r^{yU}- y_{ro}) \quad \Leftrightarrow \quad [(1+p_{ro}) y_{ro}\leq M_r^{yU}]\quad \forall r \in R^O	
\label{Eq:36}
 \end{equation}  
\begin{equation} \label{Eq:37}
\Pi_o, Q_o  , P_o  \ge 0	
 \end{equation}
\indent \textbf{The primal program aims to have the total virtual gap (TVG):}  
 \begin{equation}
\begin{aligned}
	\Delta_o^{ohPT\star}& (\$)= \max_{ V_o, U_o,  D_o^x,  D_o^y}\quad 
    \lbrace\sum_{\forall i\in I^C} v_{io} x_{io}
    +\sum_{\forall i\in I^O} [v_{io} x_{io}+(M_i^{xL}-x_{io}) d^x_{io}]\rbrace\\
    &- \lbrace\sum_{\forall r\in R^C} u_{ro} y_{ro} +\sum_{\forall r\in R^O} [u_{ro} y_{ro}+(M_r^{yU}-y_{ro}) d^y_{ro}]\rbrace
      \quad \forall o \in \mathcal{E}^{owPT}	
 \label{Eq:38}
 \end{aligned}
 \end{equation}
 \quad \quad subject to:
 \begin{equation}\begin{aligned}
     \quad \pi_{oj}:  \sum_{\forall i\in I^C \cup I^O} v_{io} x_{ij}  - \sum_{\forall r\in R^C \cup R^O}u_{ro}  y_{rj} & \leq 0(\$) \quad\forall j\in  \mathcal{E}^{owPT} \neq o \\
     \Leftrightarrow (\Delta_{oj}^{ohPT} &\leq \$0)\quad \label{Eq:39}
 \end{aligned}\end{equation}
 \begin{equation}
	q_{io}: x_{io} v_{io} \leq\tau_o(\$)\quad \forall i\in I^C	
 \label{Eq:40}
 \end{equation} 
 \begin{equation}
	q_{io}: x_{io} (v_{io}-d_{io}^x) \leq\tau_o(\$)\quad \forall i\in I^O	
 \label{Eq:41}
 \end{equation}
\begin{equation}\label{Eq:42}
	p_{ro}: y_{ro} u_{ro}\leq \tau_o(\$)\quad \forall r \in R^C \end{equation}\begin{equation}\label{Eq:43}
	p_{ro}: y_{ro} (u_{ro}- d_{ro}^y) \leq \tau_o(\$)\quad \forall r \in R^O
 \end{equation}
\begin{equation}
	 V_o, U_o, D_o^x, D_o^y \ge 0	
 \label{Eq:44}
 \end{equation}

To reduce complexity, the comprehensive examination similar to Equation (\ref{Eq:14}) is intentionally excluded. The importance of $\alpha_o^\star$ and $\beta_o^\star$ is reversed. As evidenced in Equation (\ref{Eq:45}), $DMU_o$ represents a negative hyper-virtual gap. 

\begin{equation} \label{Eq:45}
\begin{aligned}
&\$0 \leq \Delta_{o}^{ohPT\#}(\$) =(vInput^\#)-(vOutput^\#) = \alpha_o^\# - \beta_o^\#\\
&\quad=[V_o^\# x_o +\sum_{\forall i\in I^O} (M_i^{xL}-x_{io}) d_{io}^{x\#}] -[U_o^\# y_o +\sum_{\forall r\in R^O} (M_r^{yU}-y_{ro}) d_{ro}^{y\#}] \\
&\$0 \leq \Delta_{o}^{ohPT\star }(\$)=(vInput^\star)-(vOutput^\star)=\alpha_o^\star -\beta_o^\star\\
&\quad=[V_o^\star x_o +\sum_{\forall i\in I^O} (M_i^{xL}-x_{io}) d_{io}^{x\star}] -[U_o^\star y_o +\sum_{\forall r\in R^O} (M_r^{yU}-y_{ro}) d_{ro}^{y\star}] <\$1
\end{aligned}
 \end{equation}
Equation (\ref{Eq:46}) consolidates the solutions from Equation (\ref{Eq:45}) for each $DMU_j$ during Steps I and II into pairs labeled as (\textit{vInput, vOutput}), specifically $(\alpha_{oj}^\#, \beta_{oj}^\#)$ and $(\alpha_{oj}^\star, \beta_{oj}^\star)$. In the course of Step I, the value of $\Delta_{oj}^{ohPT\#}$ is found to be potentially less than -$\$\infty$. 
\begin{equation} \label{Eq:46}
\begin{aligned}
& \Delta_{oj}^{ohPT\#} (\$)= V_o^\# x_j -U_o^\# y_j = \alpha_{oj}^\# - \beta_{oj}^\# \leq \$0\quad \forall j \in \mathcal{E}^{owPT} \neq o\\
 & \Delta_{oj}^{ohPT\star} (\$)= V_o^\star x_j - U_o^\star y_j = \alpha_{oj}^\star -\beta_{oj}^\star \leq \$0 \quad \forall j \in \mathcal{E}^{owPT} \neq o
\end{aligned}
 \end{equation}
\subsection {Determine a Unified Goal Price of \texorpdfstring{$DMU_o$}.} \label{sec:5.2}
$DMU_o$ needs to ensure $\delta_o^{ohPT\star}$ and $\Delta_o^{ohPT\star}$ in Step II are within the range of [$\$0, \$1$).

The results from Steps I and II exhibit these correlations.
\begin{equation} \begin{aligned}\label{Eq:47}
\tau_o^\# :\tau_o^\star   =
 1 :  \bar{t} &=  \delta_o^{ohPT\#}:\delta_o^{ohPT\star}  =\Delta_o^{ohPT\#}:\Delta_o^{ohPT\star}\\
 &= (\alpha_o^\#- \beta_o^\# ) : (\alpha_o^\star-\beta_o^\star)
 \end{aligned}\end{equation}
Consequently, the subsequent \textit{relational equation} holds true.
\begin{equation} \label{Eq:48}
\bar{t} \$ (\alpha_o^\#- \beta_o^\# ) = 1\$(\alpha_o^\star-\beta_o^\star) 
\end{equation}
We employed Equation(\ref{Eq:49}) to calculate the dimensionless quantity $\bar{t}$. 
\begin{equation} \label{Eq:49}
	\$\bar{t} = \$1/\alpha_o^\# \textrm {  and  } \tau_o^\star   = \$ \bar{t} 	
  \end{equation}
Equation (\ref{Eq:49}) can be reformulated as Equation (\ref{Eq:50}).
\begin{equation}\begin{aligned} \label{Eq:50} 
\$(1- \beta_o^\#/\alpha_o^\# ) = 1\$(\alpha_o^\star-\beta_o^\star) \textrm{ or }
\$(1- \beta_o^\#/\alpha_o^\# )= \$(1-\beta_o^\star/\alpha_o^\star)
 \end{aligned}\end{equation}
  By dividing Equation (\ref{Eq:50}) by $\alpha_o^\star$, the equation transforms into:
 \begin{equation}\label{Eq:51}
  (\$1/\alpha_o^\star)( 1-\beta_o^\#/ \alpha_o^\#) =  \$(1-\beta_o^\star/\alpha_o^\star)
  \end{equation}
  Equation (\ref{Eq:51}) indicates that $ \$0 \leq (\alpha_o^\# - \beta_o^\# )$ and $ \$0 \leq (\alpha_o^\star - \beta_o^\star) < \$1 $, leading us to the following relational equations:
  \begin{equation} \label{Eq:52}
  0\leq( 1-\beta_o^\#/ \alpha_o^\#) \text{ and } 0\leq (1-\beta_o^\star/\alpha_o^\star)< 1
  \end{equation}
When $\alpha_o^\star$ is set to \$1 in Equation (\ref{Eq:52}), it leads to $\beta_o^\star$ being less than \$1, thereby establishing that Equation (\ref{Eq:51}) lies within the interval [0, 1). It is anticipated that $\Delta_o^{ohPT\star}$ will also be bounded in the interval [$0, \$1$). Equation (\ref{Eq:51}) indicates that the virtual inefficiency score for $DMU_o$ is independent of the unified goal price. For determining the virtual input and virtual output for $DMU_j$, refer to Equation (\ref{Eq:53}) and utilize the subsequent equations.
\begin{equation}
\begin{aligned}\label{Eq:53} 
	1 \leq F_o^{ohPT\#}&=F_o^{ohPT\star}  =\beta_o^\#/\alpha_o^\#= \beta_o^\star/\alpha_o^\star \\
    1 \leq F_{oj}^{ohPT\#}&=F_{oj}^{ohPT\star}  =\beta_{oj}^\#/\alpha_{oj}^\#= \beta_{oj}^\star/\alpha_{oj}^\star  \quad \forall j \in \mathcal{E}^{owPT} \neq o
\end{aligned}
\end{equation}
 In this analysis, the emphasis lies on assessing the virtual gap instead of the hyper-inefficiency score itself. The virtual gap for $DMU_o$, given by \(\Delta_{o}^{ohPT\star} = \alpha_o^\star - \beta_o^\star\), falls within [$\$0, \$1$).

The optimization process determines intensity levels $\pi_{oj}^{ohPT\star}$ for each $DMU_j$. For $DMU_o$, \textbf{reference peers} are in sets $\mathcal{E}_o^{ohPT}$, with $\pi_{oj}^{ohPT\star}>0$ and $\Delta_{oj}^{ohPT\star}=0$. These peers are less efficient among \textit{n} DMUs but outperform $DMU_o$, with a positive hypo-virtual gap $\Delta_o^{ohPT\star}>0$. Thus, $DMU_o$ must reduce inputs and increase outputs according to estimated rates to match or exceed its peers.

Conversely, if $\Delta_{oj}^{ohPT\star}>0$ and $\pi_{oj}^{ohPT\star}=0$, $DMU_j$ are excluded from $\mathcal{E}_o^{ohPT}$. This normalization process is demonstrated by Equation (\ref{Eq:54}), which is comparable to what is shown in Equation (\ref{Eq:24}).
\begin{equation} \begin{aligned}
 (  Q_o^\star, P_o^\star,  \Pi_o^\star)=& ( Q_o^\#, P_o^\#,  \Pi_o^\#)\\
  (\tau_o^\star,  \Delta_o^{ohPT\star },  V_o^\star, U_o^\star, D_o^{x\star}, D_o^{y\star})
  =&\bar{t} (\tau_o^\#,\Delta_o^{ohPT\#},V_o^\#, U_o^\#,  D_o^{x\#}, D_o^{y\#})
\label{Eq:54} \end{aligned}
\end{equation}
 This allows for the evaluation of the \textit{n} assessments of various $DMU_o$ through the use of normalized results, facilitating comparisons.

\subsection{Attributes of Dualities}\label{sec:5.3}
The ohPT framework shares two primary traits with the owPT framework, but it stands apart by having its input and output metrics switched. The conditions for the TAP program are specified in Equations (\ref{Eq:55}) to (\ref{Eq:57}).
\begin{equation} \begin{aligned} \label{Eq:55}
\relax [\sum_{\forall j\in \mathcal{E}^{owPT} \neq o} x_{ij} \pi_{oj}^\star - x_{io} (1 - q_{io}^\star)] \times  v_{io}^\star =\$0\quad
\forall i\in I^C \cup I^O  
\end{aligned}  \end{equation}
\begin{equation} \begin{aligned}
&\relax [\sum_{\forall j\in \mathcal{E}^{owPT} \neq o} y_{rj}\pi_{oj}^\star - y_{ro}(1+p_{ro}^\star )] \times u_{ro}^\star =\$0 \quad \forall r\in R^C \cup R^O
 \label{Eq:56} \end{aligned} \end{equation}
 \begin{equation} \begin{aligned}\label{Eq:57}
    [(1-q_{io}^\star) x_{io}-M_i^{xL}]  \times d_{io}^{x\star}&=\$0\quad \forall i \in I^O\\ [M_r^{yU}-(1+p_{ro}^\star)y_{ro}]  \times d_{ro}^{y\star}&=\$0\quad \forall r \in R^O  
\end{aligned}\end{equation} 
\indent TVP program has the conditions as shown in Equation (\ref{Eq:58}) and Equation (\ref{Eq:59}).
\begin{equation}
	(V_o^\star x_j -U_o^\star y_j ) \times \pi_{oj}^\star = \$0\quad \forall j\in \mathcal{E}^{owPT} 
	\label{Eq:58}
\end{equation} 
\begin{equation}\begin{aligned}\label{Eq:59} 
	(v_{io}^\star x_{io} - \tau_o^\star)  \times q_{io}^\star &= \$0\quad \forall i \in I^C\\\quad [(v_{io}^\star -d_{io}^{x\star}) x_{io} - \tau_o^\star ]  \times q_{io}^\star &= \$0\quad \forall i \in I^O \\
    (u_{ro}^\star y_{ro} - \tau_o^\star)  \times p_{ro}^\star&= \$0\quad \forall r\in R^C\\\quad [(u_{ro}^\star-d_{ro}^{y\star}) y_{ro} - \tau_o^\star ] \times p_{ro}^\star&= \$0\quad \forall r\in R^O 
   \end{aligned} \end{equation}

The above SCSC equations must be processed by Step-II computations. To determine Step I conditions, replace each decision variable's superscript '$\star$' with '$\#$' and substitute $\tau_o^\star$ with $\$1$. Equation (\ref{Eq:54}) outlines the connections between the solutions of the two steps.

In Equation (\ref{Eq:57}), when the ordinal data $y_{ro}$ is calibrated between two neighboring Likert points, such that $(1+p_{ro}^\star) y_{ro}<M_r^{yU}$, we find that $d_{ro}^{y\star}=0$ and the virtual unit price becomes $u_{ro}^\star$. Conversely, if $y_{ro}$ reaches the uppermost point, where $(1+p_{ro}^\star) y_{ro}=M_r^{yU}$, the virtual unit price changes to $u_{ro}^\star-d_{ro}^{y\star}>0.$ This leads to a penalty price $(M_r^{yU}-y_{ro})d_{ro}^{y\star}$ as shown in Equation (\ref{Eq:45}). The explanations related to ordinal data $x_{io}$ are not reiterated.

\section{Examination of the Derived Optimal Solutions}\label{sec:6} 

\subsection{Ranking the DMUs} \label{sec:6.1} In Stage I, the owPT model determines the reference group for $DMU_o$, denoted as $\mathcal{E}^{owPT}_o$, which is composed of DMUs possessing input-output structures similar to $DMU_o$. The set of leading DMUs that are part of $\mathcal{E}^{obPT}$ via Equation (\ref{Eq:25}) may not be uniform. 

In Stage I, the non-worst DMUs are grouped into the set $\{J-\mathcal{E}^{owPT}\}$. Within this group, $DMU_{(1)}$ is identified as the most effective and is ranked in the \textit{1st} position, having the largest virtual gap, $\Delta_{(1)}^{owPT\star}>\$0$.

In Stage II, the ohPT model determines a positive hypo-virtual gap for each $DMU_o$ within $\mathcal{E}^{owPT}$, denoted as $\delta_o^{ohPT\star}(=\Delta_o^{ohPT\star})$, which represents the improvement threshold when inputs and outputs are adjusted by the factors $Q_o^{\star}$ and $P_o^{\star}$. $DMU_{(n)}$ is recognized as the worst-performing unit, inferior than others with the largest negative hypo-virtual gap, $\Delta_{(n)}^{ohPT\star}>\$0$.

According to Equation (\ref{Eq:60}) regarding worst practices, $DMU_{(j)}$ exceeds $DMU_{(j+1)}$ in Stage I and Stage II.
\begin{equation} \begin{aligned} \label{Eq:60}
   DMU_{(j)}\succ DMU_{(j+1)} \mid \Delta_{(j)}^{owPT\star}\geq \Delta_{(j+1)}^{owPT\star}>\$0 & \quad \forall [(j),(j+1)] \in J-\mathcal{E}^{owPT} \\   
     DMU_{(j)}\succ DMU_{(j+1)} \mid \Delta_{(j+1)}^{ohPT\star}\geq \Delta_{(j)}^{ohPT\star}>\$0 & \quad \forall [(j),(j+1)] \in \mathcal{E}^{owPT}    
\end{aligned} \end{equation}

The owPT and ohPT Ord-VGA-models exhibit a trait known as \textit{unit invariance}, signifying that choosing different measurement units (e.g., tons versus kilograms) does not impact the evaluation results. This capability aids in achieving consistent assessments across distinct scales. The upcoming sections explore the key benefits of analyzing the solutions generated by the owPT and ohPT models.

\subsection{Calculating the Target Inputs and Outputs for Each Performance Measure} \label{sec:6.2} $DMU_o$ determines the target for each $X_i$ and $Y_r$ in the owPT and ohPT models, using Equations (\ref{Eq:61}) and  (\ref{Eq:62}). The procedure entails $DMU_o$ duplicating its benchmark peers using the calculated intensities $\pi_{oj}^\star$ for every $j\in \mathcal{E}_o^{owPT}$ and $j\in \mathcal{E}_o^{ohPT}$, respectively.

\begin{equation}\label{Eq:61}
\begin{aligned}
	&\widehat{x}_{io}^{owPT} = \sum_{\forall j\in \mathcal{E}_o^{owPT}} x_{ij}  \pi_{oj}^\star  = x_{io} (1 + q_{io}^\star )\quad \forall i\in I^C \cup I^O \\
	&\widehat{y}_{ro} ^{owPT} = \sum_{\forall j\in \mathcal{E}_o^{owPT}}y_{rj} \pi_{oj}^\star =y_{ro} (1 - p_{ro}^\star  )\quad  
 \forall r\in R^C \cup R^O
\end{aligned}
\end{equation}

\begin{equation}\label{Eq:62}
\begin{aligned}
 &\widehat{x}_{io}^{ohPT} = \sum_{\forall j\in \mathcal{E}_o^{ohPT}} x_{ij}  \pi_{oj}^\star  = x_{io} (1 - q_{io}^\star )\quad \forall i\in I^C \cup I^O \\
	&\widehat{y}_{ro} ^{ohPT} = \sum_{\forall j\in \mathcal{E}_o^{ohPT}}y_{rj} \pi_{oj}^\star =y_{ro} (1 + p_{ro}^\star  )\quad  
 \forall r\in R^C \cup R^O
\end{aligned}
\end{equation}

\subsection{Validate the Solutions}\label{sec:6.3}
In the owPT and ohPT models, the altered $X_i$ and $Y_r$ for $DMU_o$, as specified by Equation (\ref{Eq:63}), must satisfy the following conditions:
\begin{equation}\begin{aligned}\label{Eq:63}
\widehat{x}_{io}^{owPT} \leq M_{io}^{xU}\quad \forall i \in I^O,\quad & \widehat{y}_{ro}^{owPT} \geq M_{ro}^{yL}\quad \forall r \in R^O \\
\widehat{x}_{io}^{ohPT} \geq M_{io}^{xL}\quad \forall i \in I^O, \quad  &\widehat{y}_{ro}^{ohPT} \leq M_{ro}^{yU}\quad  \forall r \in R^O\\
\end{aligned}
\end{equation}

Employing Equations (\ref{Eq:61}) and (\ref{Eq:62}) for the owPT and ohPT models respectively, the \textit{benchmark virtual scales} (bvInput, bvOutput) are calculated with Equation (\ref{Eq:64}). This method of deducing and refining the benchmarks demonstrates the adjustment or calibration of initial estimates from the two models, thus reducing the span of $\Delta_o^{owPT\star}$ and $\Delta_o^{ohPT\star}$.
\begin{equation} \label{Eq:64} 
\begin{aligned}
	\widehat{\alpha}_o^w= \sum_{\forall i\in I^C \cup I^O} \widehat{x}_{io}^{owPT} v_{io}^\star, \quad &\widehat{\beta}_o^w= \sum_{\forall r\in R^C \cup R^O}\widehat{y}_{ro}^{owPT}  u_{ro}^\star\\
    \widehat{\alpha}_o^h= \sum_{\forall i\in I^C \cup I^O} \widehat{x}_{io}^{ohPT}  v_{io}^\star, \quad &\widehat{\beta}_o^h= \sum_{\forall r\in R^C \cup R^O}\widehat{y}_{ro}^{ohPT}  u_{ro}^\star	
    \end{aligned}
	\end{equation}	
Moreover, the coordinates ($\widehat\alpha_o^w$, $\widehat\beta_o^w$) and ($\widehat\alpha_o^h$, $\widehat\beta_o^h$) are positioned at the prime meridian and equator in the two-dimensional structures of the owPT and ohPT solutions, as depicted in figures within Section \ref{sec:7}. Importantly, $\widehat\alpha_o^w=\widehat\beta_o^w$ and $\widehat\alpha_o^h=\widehat\beta_o^h$ constitute the fundamental conditions necessary to accurately evaluate the owPT and ohPT models.

If $DMU_o$ displays no virtual gap, both its inputs and outputs remain constant. 

\subsection{Visualize the Virtual Technology Sets in 2D Graphics} \label{sec:6.4} Equation (\ref{Eq:66}) characterizes the group of \textit{virtual technologies} within the context of the owPT model when evaluating $DMU_o$. Equations (\ref{Eq:15}) and (\ref{Eq:16}) define the virtual scale for each $DMU_j$ in terms of \textit{vInput} and \textit{vOutput}. During Step II, these virtual scales for each $DMU_j$ illustrate specific aspects or features within the model. As depicted in the two-dimensional graphical intuitions in Section \ref{sec:7}, all points reside above or on the prime meridian. The virtual output for $DMU_o$ is standardized as $\beta_o^{\star}$, with a value of \$1, thus $DMU_o$ identifies its unique $\Phi_o^{owPT\star}$. 
\begin{equation}
    \Phi_o^{owPT\star}=\lbrace (\alpha_o^{w\star}, \beta_o^{w\star}) \mid (\alpha_j^{\star},\beta_j^{\star})\quad \forall j\in J   \rbrace \cup (\widehat\alpha_o^w, \widehat\beta_o^w)
	\label{Eq:66}
	\end{equation}
   
  The particular set of virtual technologies $\Phi_o^{ohPT\star}$ defined as Equation (\ref{Eq:67}) is assigned to $DMU_o$ in the ohPT evaluation. In Step II, $DMU_o$ is evaluated and $(\alpha_j^{\star},\beta_j^{\star})$ is ascertained, as referenced in Equations (\ref{Eq:45}) and (\ref{Eq:46}). The virtual input of $DMU_o$ undergoes normalization to $\alpha_o^{\star}$, resulting in an equivalence of \$1. 
    \begin{equation}
    \Phi_o^{ohPT\star}=\lbrace (\alpha_o^{h\star}, \beta_o^{h\star}) \mid (\alpha_j^{\star},\beta_j^{\star})\quad \forall j\in \mathcal{E}_o^{owPT}\rbrace \cup (\widehat\alpha_o^h, \widehat\beta_o^h)
	\label{Eq:67}
	\end{equation}
The two-dimensional graphical intuitions of the two numerical examples, as shown in the figures in Section \ref{sec:7} that all points are located above or on the equator while $DMU_o$ is located under the Equator.

\section{Numerical Examples} \label{sec:7} The minimal working example Table \ref{table1} and the large scale real problem \cite{Chen} in appendix are used to illustrate the innovative MCA method. The yielded optimal results as detailed in Tables \ref{table2} and \ref{table3}. These solutions facilitate an examination of the properties of the owPT and ohTP models.

\subsection{Minimal Working Example}\label{sec:7.1}
 Table \ref{table2} provides comprehensive solutions for assessing the six Decision Making Units (DMUs) delineated in Table \ref{table1}. In Tables \ref{table2}, each $DMU_o$ determines a unified goal price, $\tau_o^\star$, to be the lower and upper bounds of virtual prices of input and output metrics in the owPT and ohPT models, respectively. The model optimizes objective values alongside primal and dual decision variables are concurrently illustrated.

 \subsubsection{Stage I analysis.} \label{sec:7.1.1}
 
 In row R1, the respective columns K, B, D, G, and H have the virtual gaps are zero, that $DMU_K$, $DMU_B$, $DMU_D$, $DMU_G$, and $DMU_H$ cannot be compared. These five DMUs are the bottom-tier. In row R3, the calculated intensity values $\pi_{oK}^\star$, $\pi_{oB}^\star$, $\pi_{oD}^\star$, $\pi_{oG}^\star$, and $\pi_{oH}^\star$, found in columns K, B, D, G, and H, respectively, are all equal to 1, while the other intensities are zero. This reveals that the input-output configurations of the other DMUs differ significantly from $DMU_o$. 
 
\begin{figure}[ht] \centering
 \includegraphics
 [width=0.65\linewidth, height=3 in] {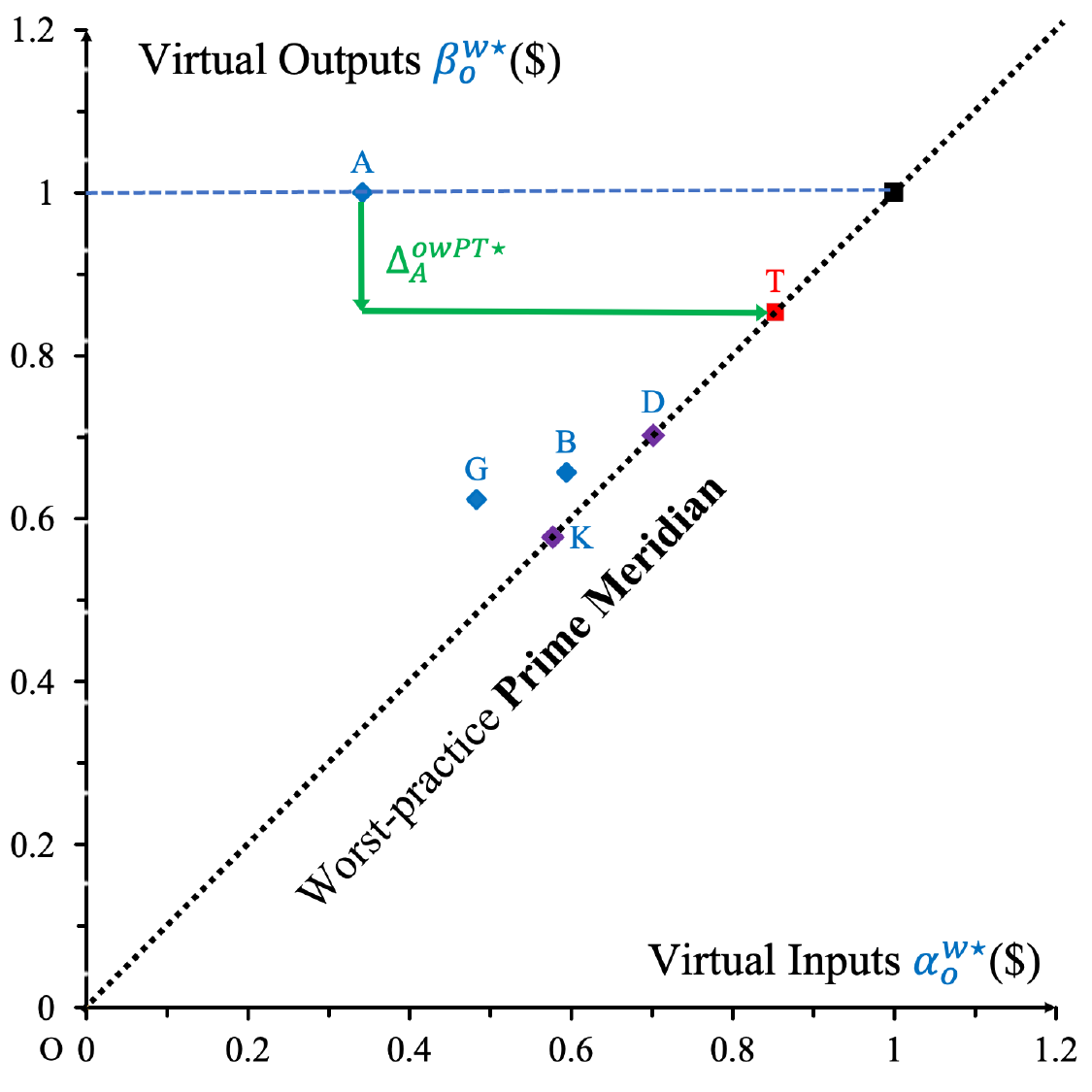} 
 \caption {Assessment solutions of DMU-A in Stage I} \label{fig3}
  \end{figure}

 The outcomes pertaining to $DMU_o$(=$DMU_A$) in Stage I are depicted in Figure \ref{fig3}, which presents a two-dimensional visualization of the virtual gap analysis. The points T, K, and D are correspond to the locations of the target of $DMU_A$, $DMU_K$, and $DMU_D$, respectively. Each of these points is situated on the \textit{worst-practice}\textbf{ prime meridian}, indicating a congruence between the virtual input and virtual output. $DMU_o$ intersects the vertical axis at the coordinate (1,0).
Unlike the others, $DMU_B$ and $DMU_G$ exhibit virtual gaps exceeding zero, which positions points B and G above the prime meridian. The relevant data for $DMU_A$ can be found in its respective column in Table \ref{table2}.

\begin{figure}[ht] \centering
 \includegraphics
 [width=0.63\linewidth, height=4.6 in] {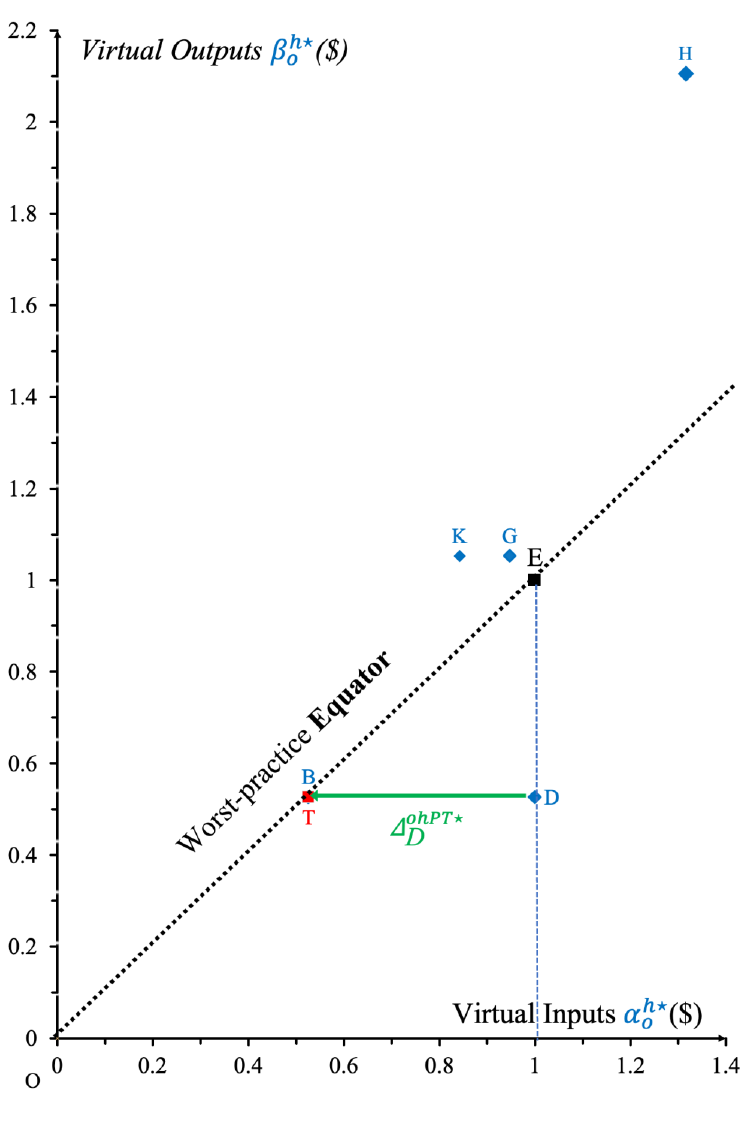} 
 \caption {Assessment solutions of DMU-D in Stage II} \label{fig4}
  \end{figure}  
  
\begin{table}[!htp]
\centering
\setlength{\tabcolsep}{0.64 pt}
\renewcommand{\arraystretch}{1}
\caption{Evaluation Solutions in Stages I and II of DMUs in Table \ref{table1}} \label{table2}

\small \begin{tabular*}
{\textwidth}{@{\extracolsep\fill}lccrrrrrrcrrrrrr}  \hline
Row&Sym-&Metric&\multicolumn{6}{@{}c@{}}{$DMU_o$ in Stage I owPT model}&&\multicolumn{5}{@{}c@{}}{$DMU_o$ in Stage II ohPT model} \\\cmidrule{4-9} \cmidrule{11-15}
	&	bol&	Unit&	K	&	A	&	B	&	D	&	G	&	H	&	&	K	&	B	&	\textbf{D}$\star$	&	G	&	H	\\ \cmidrule{1-3}   \cmidrule{1-9} \cmidrule{11-15}
R1& $\tau_o^\star$	& $\$$	&	0.500	&	0.447	&	0.222	&	0.033		&	0.036	&	0.017	&	&	0.639	&	1	&	1	&	1	&	1	\\
& $\Delta_o^\star, \delta_o^\star$ & $\$$ &	0	&	\textbf{0.6}	&	0	&		0	&	0	&	0	&	&	0.361	&	0.222	&	\textbf{0.474}$\star$	&	0.044	&	0.086	\\ 
\cmidrule{1-9} \cmidrule{11-15}
R2& $v_{1o}^\star$ & $\$$/kg	&0.313	&	0.194	&	0.222	&	0.483		&	0.535	&	0.393	&	&	0.243	&	0	&	0.526	&	0.556	&	0.4	\\
& $v_{2o}^\star$ & $\$$/(pt.)&	0.125	&	0.067	&	0.130	&	0.016		&	0.018	&	0.017	&	&	0.153	&	0.167	&	0	&	0	&	0	\\
& $u_{1o}^\star$ & $\$$/(lvl.) &	0.250	&	0.149	&	0.223	&	0.223		&	0.247	&	0.004	&	&	0	&	0.164	&	0.526	&	0.2379	&	0	\\
& $u_{2o}^\star$ & $\$$/(piece)	&	0.010	&	0.006	&	0.009	&	0.008		&	0.009	&	0.014	&	&	0.013	&	0.007	&	0	&	0.0084	&	0.0131	\\
& $d_{2o}^{x\star}$ & $\$$/(pt.)	&	0	&	\textbf{0.082}	&	0	&	0	&		0	&	0	&	&	0	&	0	&	0	&	0	&	0	\\
& $d_{1o}^{y\star}$ & $\$$/(lvl.)	&0	&	0	&	0	&	0	&	0	&	0	&	&	0	&	0	&	0	&	0	&	0	\\ \cmidrule{1-9}\cmidrule{11-15}
R3& $q_{1o}^\star$ &-&	0	&	0.015	&	4E-16	&	4E-16	&		0	&	0	&	&	0	&	0	&	0.474	&	0.044	&	0.086	\\
&  $q_{2o}^\star$ &-&	0	&	1	&	0	&	4E-16&		&	2E-16		&	&	0	&	0.222	&	0	&	0	&	0	\\
& $p_{1o}^\star$ &-&		0	&	0.329	&	0	&	0		&	0	&	0	&	&	0	&	0	&	0	&	0	&	0	\\
& $p_{2o}^\star$ &-&		0	&	0	&	0	&	0		&	0	&	0	&	&	0.566	&	0	&	0	&	0	&	0	\\
& $\pi_{oK}^\star$ &-&	1	&	0.678	&	5E-17	&	1E-16		&	3E-16	&	0	&	&	-	&	0.055	&	0	&	0.945	&	1.429	\\
& $\pi_{oA}^\star$ &-&	0	&	0	&	0	&	0		&	0	&	0	&	&	-	&	-	&	-	&	-	&	-	\\
& $\pi_{oB}^\star$ &-&	0	&	0	&	1	&	0		&	0	&	0	&	&	0.455	&	0.890	&	1	&	0	&	0	\\
& $\pi_{oD}^\star$ &-&	0	&	0.657	&	0	&	1		&	1E-15	&	0	&	&	0	&	0	&	0	&	0.110	&	0	\\
& $\pi_{oG}^\star$ &-&	0	&	0	&	0	&	0	&	1	&	0	&	&	0.636	&	0	&	0	&	-	&	0	\\
& $\pi_{oH}^\star$ &-&	0	&	0	&	0	&	0		&	0	&	1	&	&	0	&	0	&	0	&	0	&	-	\\ \cmidrule{1-9} \cmidrule{11-15}
R4& $v_{1o}^\star x_{1o}$ & $\$$&	0.5	&	0.447	&	0.222	&	0.918	&	0.964	&	0.983	&	&	0.389	&	0	&	1	&	1	&	1	\\
\multicolumn{3}{l}{($v_{2o}^\star\pm d_{2o}^{x\star}) x_{2o}\quad \$$}&	0.5	&	0.940	&	0.778	&	0.082		&	0.036	&	0.017	&	&	0.611	&	1	&	0	&	0	&	0	\\
\multicolumn{3}{l}{($u_{1o}^\star\pm d_{1o}^{y\star}) y_{1o}\quad \$$}&	0.5	&	0.447	&	0.223	&	0.223		&	0.494	&	0.017	&	&	0	&	0.164	&	0.526	&	0.476	&	0	\\
&$u_{2o}^\star y_{2o}$&$\$$&	0.5	&	0.553	&	0.777	&	0.777&	0.506	&	0.983	&	&	0.639	&	0.614	&	0	&	0.480	&	0.914	\\ 
\cmidrule{1-9} \cmidrule{11-15}
R5& $\widehat\alpha_o^\star, \widehat\beta_o^\star$ & $\$$&	1	&	0.853	&	1	&	1	&	1	&	1	&	&	1	&	0.778	&	0.526	&	0.956	&	0.914	\\ 
\cmidrule{1-9} \cmidrule{11-15}
R6& $\alpha_{oK}^\star$ & $\$$ &	1	&	0.577	&	0.874	&	0.839	&	0.929	&	0.697	&	&	1	&	0.667	&	0.842	&	0.889	&	0.640	\\
& $\alpha_{oA}^\star$ & $\$$ &	1.094	&	0.342	&	0.899	&	1.161	&	1.286	&	0.955	&	&	-	&	-	&	-	&	-	&	-	\\
& $\alpha_{oB}^\star$ & $\$$ &	1.063	&	0.594	&	1	&	0.581	&	0.644	&	0.494	&	&	1.160	&	1	&	0.526	&	0.556	&	0.400	\\
& $\alpha_{oD}^\star$ & $\$$ &	1.219	&	0.702	&	1.070	&	1		&	1.108	&	0.831	&	&	1.226	&	0.833	&	1	&	1.056	&	0.760	\\
&$ \alpha_{oG}^\star$ & $\$$ &	0.813	&	0.483	&	0.659	&	0.903	&	1	&	0.742	&	&	0.743	&	0.333	&	0.947	&	1.000	&	0.720	\\
& $\alpha_{oH}^\star$ & $\$$ &	0.906	&	0.552	&	0.684	&	1.225	&	1.357	&	1	&	&	0.760	&	0.167	&	1.316	&	1.389	&	1.000	\\ 
\cmidrule{1-9} \cmidrule{11-15}
R7& $\beta_{oK}^\star$ & $\$$ &	1	&	0.577	&	0.874	&	0.839	&	0.929	&	0.697	&	&	0.639	&	0.667	&	1.053	&	0.889	&	0.640	\\
& $\beta_{oA}^\star$ & $\$$ &1.740	&	1	&	1.516	&1.446	&	1.602	&	1.375	&	&	-	&	-	&	-	&	-	&	-	\\
& $\beta_{oB}^\star$ & $\$$ &	1.158	&	0.657	&	1	&	0.936	&	1.037	&	1.254	&	&	1.160	&	0.778	&	0.526	&	0.988	&	1.162	\\
& $\beta_{oD}^\star$ & $\$$ &	1.240	&	0.702	&	1.070	&	1	&	1.108	&	1.367	&	&	1.264	&	0.833	&	0.526	&	1.056	&	1.267	\\
& $\beta_{oG}^\star$ & $\$$ &	1.082	&	0.623	&	0.943	&	0.903	&	1	&	0.809	&	&	0.743	&	0.722	&	1.053	&	0.956	&	0.744	\\
&$ \beta_{oH}^\star$ & $\$$ &	1.714	&	0.995	&	1.503	&	1.453	&	1.610	&	1	&	&	0.912	&	1.140	&	2.105	&	1.542	&	0.914	\\ 
\cmidrule{1-9} \cmidrule{11-15}
 \end{tabular*}
 \end{table}

 The rectilinear distance $\Delta_A^{owPT\star}(=\$0.6)$ indicates the virtual gap price between point A and point T. The solutions are evaluated with respect to the peers $DMU_K$ and $DMU_D$. The figure visually corroborates the precision of the evaluation for $DMU_o$. 

Using the solutions in column A of Table \ref{table2}, evaluate the equation for $x_{2o}$ from the SCSC equations. (\ref{Eq:29}) and  (\ref{Eq:31}):
\begin{equation*} 
\begin{aligned}
\lbrace(1+q_{2o}^\star)x_{2o}-M_2^{xU}\rbrace \times d_{2o}^{x\star}&= [(1+1)3-6]\times \$0.082=\$0\\ 
\quad (v_{2o}^\star +d_{2o}^{x\star}) x_{2o} - \tau_o^\star &=0\quad and \quad q_{2o}^\star >0 
\end{aligned}
\end{equation*}
This indicates that $DMU_A$ experiences an increase of \textbf{$\$ 0.447$}$(=q_{2o}^\star \tau_o^\star)$ in $\Delta_A^{owPT\star}$ due to the Likert scale concerning $x_{2o}$. It is expanded from 3 to the upper Likert scale \textbf{6} (=$\widehat{x}_{2o}^\star$). $DMU_A$ does suffer from having the $x_{2o}$ value.

Section \ref{sec:6.2} outlines the validation process for the target $DMU_o$, which ensures that the virtual input $\widehat\alpha_o^\star$ matches the virtual output $\widehat\beta_o^\star$, see row R5.

Rows R6 and R7 enumerate the virtual input and output for each $DMU_j$ concerning $DMU_o$, identified by $\alpha_{oj}^\star$ and $\beta_{oj}^\star$. The units of these symbols are aligned to maintain consistency between variables and constraints in both the primal and dual formulations.

  \subsubsection{Stage II analysis.} \label{sec:7.1.2} In the course of Stage II, as detailed in Table \ref{table2}, the ohPT model is employed to evaluate the hyper-virtual gap for $DMU_K, DMU_B$, $DMU_D$, $DMU_G$, and $DMU_H$. The corresponding outputs, as presented in row R1 of Table \ref{table2}, are 0.361, 0.222, \textbf{0.474}$\star$, 0.044, and 0.086, respectively. According to Equation (\ref{Eq:60}), $DMU_D$ is identified as the least favorable DMU concerning the MCA problem.

Figure \ref{fig4} indicates that $DMU_D$ is recognized as the poorest performer, occupying the final spot in the ranking list as $DMU_{(6)}$. Position D is positioned below the equator with a substantial virtual gap denoted as $\Delta_o^{ohPT\star}$. In contrast, $DMU_B$ is located precisely on the Equator, exhibiting a zero virtual gap, marked as $\pi_{oB}^{ohPT\star}=1$, and possesses a reference set $\mathcal{E}_o^{ohPT}$=\{B\}. The other peers are positioned above the Equator, each exhibiting negative virtual gaps, as denoted by $\Delta_{oj}^{ohPT\star}=\alpha_{oj}^\star-\beta_{oj}^\star<0$.

In the initial stage, the size of the MCA decision matrix is primarily determined by the number of Decision Making Units (DMUs) involved. Generally, only a limited number of DMUs achieve to the set of worst-DMUs. In the subsequent stage, there is a marked reduction in computational demand. Ultimately, the computational speed and the maximum problem size are contingent upon the efficacy of the linear programming software employed.

Based on the virtual gaps indicated in row R1 of Table \ref{table2} and Equation (\ref{Eq:60}), the order of ranking for the six DMUs is as follows: $DMU_A \succ DMU_G \succ DMU_H \succ DMU_B \succ DMU_K \succ DMU_D$.
\subsection{The Large Scale Real Problem} \label{sec:7.2} 

\citet{Chen} assessed the energy efficiencies across 29 provinces in China, utilizing three input and three output performance indicators \citep[see][Table 5]{Chen}. The third output metric is specifically measured using the Likert scale. This situation represents a typical real-world MCA problem that has not been adequately tackled by much of the literature developed over the last fifty years. To refine the outputs, they employed three distinct models in their data analysis \citep[refer to][Models (2), (3), and (4)]{Chen}. Subsequently, they presented three solutions \citep[see][Tables 6, 7, and 8]{Chen}.

Table \ref{table3} presents our evaluations of the four DMUs, with all inputs and outputs comprehensively modified. $DMU_1$ underperformed the others. We confirm the novel MCA method is both reliable and scalable, facilitating comprehensive assessments effectively and efficiently within decision support systems.

\subsubsection{Stage I analysis.}\label{sec:7.2.1}
By utilizing the owPT model to evaluate the virtual gap of each DMUs of the 29 DMUs, we identify the set $\mathcal{E}^{owPT}=\{2,3,4,6,10,11,12,14,16,19,20,22,23,24,26,27,28,29\}$, highlighting 18 DMUs as non-worst. The remaining 11 DMUs are the worst DMUs: 1, 5, 7, 8, 9, 13, 15, 17, 18, 21, 25.  For clarity, the assessment results presented are limited to the five worst DMUs: $DMU_1$, $DMU_{10}$, $DMU_{11}$, and $DMU_{19}$ in both Stage I and Stage II, as specified in row R1 of Table \ref{table3}.

\begin{figure}[ht] \centering
 \includegraphics
 [width=0.7\linewidth, height=3 in]{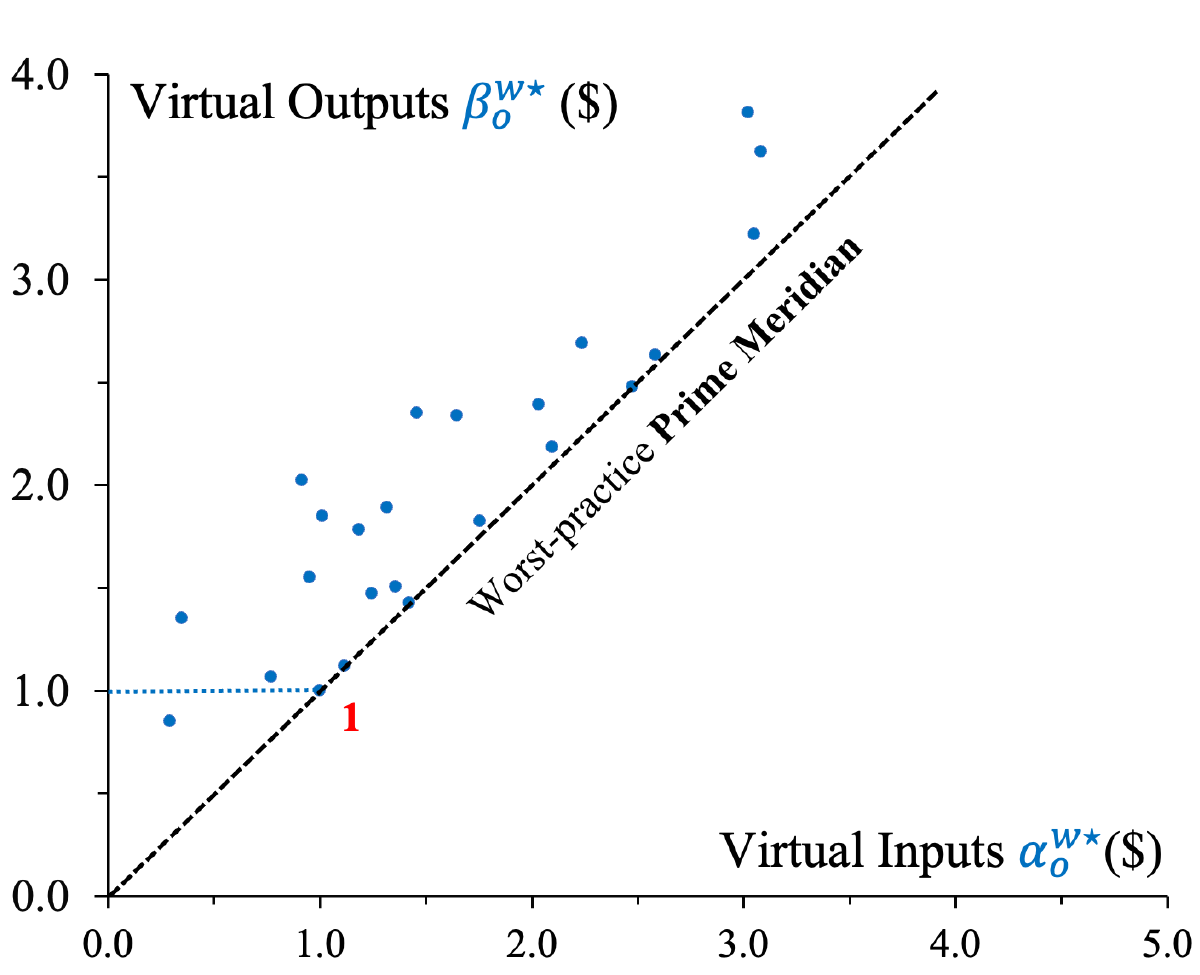} 
 \caption {Assessment solutions of DMU-1 in Stage I} \label{fig5}
  \end{figure}

For example, Figure \ref{fig5} illustrates the evaluation results for $DMU_1$, which is represented as the point "1" on the graph. It is situated on the prime meridian since its virtual gap equals 0 and $\pi_{1}^{owPT\star}$ has a value of 1. Conversely, the remaining 28 DMUs have negative virtual gaps and are located above the prime meridian. Any one of the 28 other $DMU_j$ shows a $\pi_j^{owPT\star}$ value of 0. 
   \begin{table*}[!ht]
\centering
\setlength{\tabcolsep}{2 pt}
\renewcommand{\arraystretch}{0.8}
\caption{Stage II Solutions of the ohPT VGA-models for each $DMU_o$ of \citep{Chen}} \label{table3}
 \footnotesize
\begin{tabularx}{\linewidth}{lccrrrrrrrrr}  \hline
Row&Sym-&Metric &\multicolumn{4}{c}{\textit{DMUo in Stage I}}&\quad&\multicolumn{4}{c}{\textit{DMUo in Stage II}} \\ 
&	bol&	unit&	\textbf{1}	&	\textbf{10}	&	\textbf{11}	&	\textbf{19} &&	\textbf{1}$\star$	&	\textbf{10}	&	\textbf{11}	&	\textbf{19} \\ \cmidrule{1-7} \cmidrule{9-12}
R1& $\tau_o^\star$	& $\$$	&	0.171	&	0.263	&	0.311	&	0.103	&	&	0.574	&	0.500	&	0.498	&	0.936	\\
& $\Delta_o^\star, \delta_o^\star$ & $\$$&		0E+00	&	0	&	0	&	0	&	&	\textbf{0.426}$\star$	&	0.204	&	0.113	&	0.068	\\ \cmidrule{1-7} \cmidrule{9-12}
R2& $v_{1o}^\star$ & $\$$/Rm &	1E-04	&	2E-05	&	2E-05	&	5E-06	&	&	2E-05	&	2E-05	&	2E-07	&	0E+00	\\
& $v_{2o}^\star$ & $\$$/Pn &	2E-08	&	3E-08	&	4E-08	&	8E-09	&	&	8E-08	&	0E+00	&	5E-08	&	5E-09	\\
& $v_{3o}^\star$ & $\$$/Tn &	2E-05	&	9E-06	&	2E-05	&	3E-05	&	&	4E-05	&	2E-05	&	3E-05	&	3E-05	\\
& $u_{1o}^\star$ & $\$$/Tn &	1E-05	&	6E-06	&	2E-05	&	2E-06	&	&	0E+00	&	0E+00	&	1E-06	&	0E+00	\\
& $u_{2o}^\star$ & $\$$/Rm&	6E-05	&	5E-06	&	1E-05	&	2E-05	&	&	6E-05	&	5E-06	&	1E-05	&	2E-05	\\
& $u_{3o}^\star$ & $\$$/pt. &	2E-01	&	4E-01	&	9E-02	&	1E-01	&	&	0E+00	&	5E-01	&	4E-01	&	1E-01	\\
& $d_{3o}^{y\star}$ & $\$$/pt.	&	0.000	&	0.000	&	\textbf{0.217}	&	0.000	&	&	0	&	0	&	0	&	0	\\ \cmidrule{1-7} \cmidrule{9-12}
R3& $q_{1o}^\star$	&	-&		0	&	0	&	0	&	0	&	&	0	&	0.201	&	0	&	0	\\
& $q_{2o}^\star$	&	-&	0	&	0	&	0	&	0	&	&	0.462	&	0	&	0.211	&	0	\\
&$q_{3o}^\star$	&	-&	0	&	0	&	0	&	0	&	&	0	&	0.058	&	0.015	&	0.073	\\
 &$p_{1o}^\star$	&	-&	0&		0	&	0	&	0	&&	0	&		0	&	0	& 0	\\
& $p_{2o}^\star$	&	-&	0	&	0	&	0	&	0	&	&	0.281	&	0	&	0	&	0	\\
 &$p_{3o}^\star$	&	-&	0	&	0	&	0	&	0	&&	0	&	0.150	&	0	&0	\\ \cmidrule{1-7} \cmidrule{9-12}
R4& $\pi_{o1}^\star$ &-	&	1	&	0	&	0	&	0	&	&0		&	0	&	0.462	&	0.062	\\
& $\pi_{o10}^\star$ &-	&	0	&	1	&	0	&	0	&	&0		&	0	&	0.431	&		0\\
& $\pi_{o11}^\star$ &-	&	0	&	0	&	1	&	0	&	&	0.248	&	0	&	0 	&	0.852	\\
& $\pi_{o16}^\star$ &-	&	0	&	0	&	0	&	0	&	&0&1.15		&0.043 &	0	\\
& $\pi_{o19}^\star$ &-	&	0	&	0	&	0	&	1	&	&	0.092	&	0	&	0.032	&	0	\\
& $\pi_{o23}^\star$ &-	&	0	&	0	&	0	&	0	&	&	0	&	0	&	0	&	0.543	\\ \cmidrule{1-7} \cmidrule{9-12}
R5& $v_{1o}^\star x_{1o}$ & $\$$ &	0.658	&	0.474	&	0.311	&	0.103	&	&	0.107	&	0.500	&	0.003	&	0	\\
& $v_{2o}^\star x_{2o}$ & $\$$ &	0.171	&	0.263	&	0.378	&	0.103	&	&	0.574	&	0	&	0.498	&	0.064	\\
& $v_{3o}^\star x_{3o}$ & $\$$ &	0.171	&	0.263	&	0.311	&	0.794	&	&	0.319	&	0.500	&	0.498	&	0.936	\\
& $u_{1o}^\star y_{1o}$ & $\$$	&0.220	&	0.314	&	0.567	&	0.103	&	&	0	&	0	&	0.036	&	0	\\
& $u_{2o}^\star y_{2o}$ & $\$$	&	0.585	&	0.263	&	0.340	&	0.684	&	&	0.574	&	0.296	&	0.422	&	0.702	\\
& $(u_{3o}^\star-d_{3o}^{y\star}) y_{3o}$ & $\$$	&	0.195	&	0.424	&	0.311	&	0.213	&	&	0	&	0.500	&	0.392	&	0.229	\\ \cmidrule{1-7} \cmidrule{9-12}
R7& $\widehat{\alpha}_o^\star, \widehat{\beta}_o^\star$ &$\$$	&	1	&	1	&	1	&	1	&	&	0.735	&	0.871	&	0.850	&	0.932	\\
& $\Delta_{o1}^\star$ &	$\$$ &	0 	&	0.185	&	0.000	&	0.000	&	&	\textbf{0.426}$\star$	&	-0.326	&	0	&	0\\
& $\Delta_{o10}^\star$ &	$\$$ &	0.177 	&	-0.225 	&	0.000	&	0.000	&	&	-0.966	&	\textbf{0.204}	&	0	&	-0.057\\
& $\Delta_{o11}^\star$ &	$\$$ &	0.044 	&	0.602 	&	-0.229	&	0.265	&	&	0	&	-0.069	&	\textbf{0.150}	&	0\\
& $\Delta_{o19}^\star$ &	$\$$ &	0.789 	&	0.646	&	0.104	&	-0.524	&	&	0	&	-0.420	&	0	&	\textbf{0.06} \\ \hline
\multicolumn{12}{l}{Note:
Rm=(RMB), Pn=(person), Tn(= Tons), pt.=(point of Likert scale)}
 \end{tabularx}
 \end{table*}

\begin{figure}[ht] \centering
 \includegraphics
 [width=0.7\linewidth, height=3.1 in]{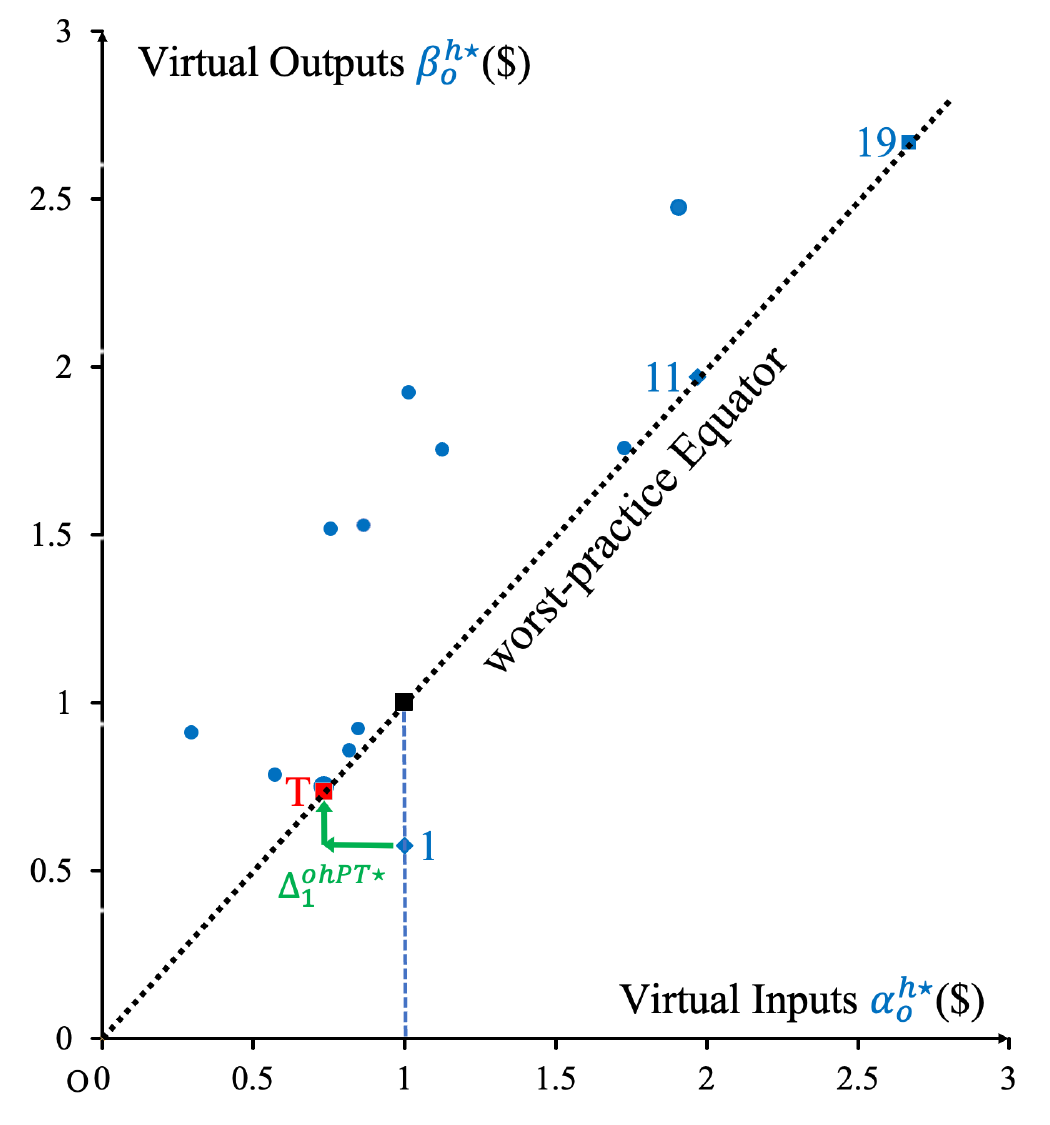} 
 \caption {Assessment solutions of DMU-1 in Stage II} \label{fig6}
  \end{figure}

 \subsubsection{ Stage II analysis.}\label{sec:7.2.2} Table \ref{table3} includes only the four DMUs that are part of $\mathcal{E}^{owPT}$. The initial row, R1, presents the hypo-virtual gaps  (\$) of 0.426, 0.204, 0.113, and 0.068. Given that $DMU_1$ has the highest hypo-gap value of \$0.426, it is identified as the weakest performer in the MCA issue. The solutions pertaining to $DMU_1$ in Stage II are recorded in the corresponding column of Table \ref{table3}. 
 
 Evaluating $DMU_1$ yields computed values for $\pi_{11}^{owPT\star}$ and $\pi_{19}^{owPT\star}$ as 0.248 and 0.092, respectively, revealing that the reference set is $\mathcal{E}_1^{ohPT}$=\{11, 19\}. Figure \ref{fig6} shows $DMU_{11}$ and $DMU_{19}$ positioned on the equatorial line with no hypo-virtual gaps. In contrast, the remaining 15 DMUs are above the equator with negative hypo-virtual gaps. The $DMU_o$ being assessed, $DMU_1$, is positioned beneath the equatorial line and shows a positive hypo-virtual gap of $\Delta_1^{ohPT\star}(=\$0.426)$, necessitating enhancement to reach its target point T. 
 
Row R3 demonstrates that $DMU_1$ needs to lower $q_{2o}^\star(=0.462)$ while also raising $p_{2o}^\star(=0.281)$. The total adjustment cost $\delta_1^{ohPT\star}$ is derived as $(0.462+0.281)\times \tau_o^\star=\$0.426$, where $\tau_o^\star=\$0.574$ is indicated in Row R1. $DMU_1$ could advance towards target T, potentially achieving performance levels similar to those of its benchmark counterparts $DMU_{11}$ and $DMU_{19}$, which show equivalent minimal performance in comparison to the other 15 DMUs.

 For the 18 non-worst DMUs reviewed in Stage II, each $DMU_o$ is analyzed using a related graph. In this setup, $DMU_o$ is positioned on the vertical axis at coordinates (1, 1) for vertical movement. Equation (\ref{Eq:52}) indicates that the normalized $\alpha_o^\star$ is equal to \$1, while $\beta_o^\star$ is within the range [\$0, \$1). The normalized hypo-virtual gap, calculated as $(\alpha_o^\star-\beta_o^\star)$, also lies within [\$0, \$1). In this analysis, $DMU_1$ exhibits the largest hypo-virtual gap and is identified as the worst alternative in the MCA framework.

 According to the virtual gaps illustrated in row R1 of Table \ref{table3} and defined by Equation (\ref{Eq:60}), the ranking sequence for the four lowest-performing DMUs is: $DMU_{19} \succ DMU_{11} \succ DMU_{10} \succ DMU_1 $.
\subsection{Verify the Precise Evaluations} \label{sec:7.3} The innovative MCA approach accurately assessed DMUs via the owPT and ohPT models. Tables \ref{table2} and \ref{table3} detail the targets for each $DMU_o$, represented by $(\widehat{\alpha}_o, \widehat{\beta}_o)$. In Figures \ref{fig3}, \ref{fig4}, \ref{fig5}, and \ref{fig6}, point T denotes the target for $DMU_o$, situated on the diagonal at the coordinates (0,0) because $\widehat{\alpha}_o= \widehat{\beta}_o$.

The calculated decision variables for the owPT and ohPT models, namely $V_o^\star, U_o^\star, P_o^\star, Q_o^\star, \Pi_o^\star$, are utilized in the equations presented in Section \ref{sec:6} to generate the 2D plots. Only the exact solutions will achieve the goal of $\widehat{\alpha}_o= \widehat{\beta}_o$. 

In the 2D graph, the dimensionless straight-line distance between the $DMU_o$ points and its target denotes the \textbf{inefficiency score} of $DMU_o$, falling within the interval (0,1). This can be observed in Figures \ref{fig3}, \ref{fig4}, \ref{fig5}, and \ref{fig6}. 

\section{Discussions and Conclusions} \label{sec:8} Various Multi-criteria Analysis (MCA) techniques have been devised to address different decision-making issues, each possessing its own advantages and limitations. The extensive analysis of these techniques in the literature highlights the intricacies in selecting the most appropriate one. The growing demand for MCA solutions creates avenues for developing novel methods. In our study, we introduce a new MCA approach that implements a two-tiered virtual gap analysis (VGA) model based on linear programming, aimed at solving MCA problems with differing complexity levels. These VGA-models can be adapted to handle a wide range of MCA issues by incorporating extra conditions. For clarity and precision in methodology, we detail the fundamental strategies for tackling MCA challenges below.

\subsection{Heterogeneity of the decision matrix}\label{sec:8.1}  
Many conventional MCA techniques presume that DMUs operate under a uniform input-output framework, which complicates the identification of heterogeneity within the decision matrix. Yet, there are situations where $DMU_o$ may not successfully determine its reference set due to this diversity. Although the owPT VGA-model can calculate $\Delta_o^{owPT\star}=\$0$, this does not imply that $DMU_o$ surpasses all other DMUs. For instance, when assessing $DMU_B$ and $DMU_H$ via the owPT VGA-model, the outcomes in row R3 of Table \ref{table2} reveal $\mathcal{E}_B^{owPT} = \{B\}$ and $\mathcal{E}_H^{owPT} = \{H\}$, demonstrating their heterogeneity in relation to each other. The optimization process identified the minimized virtual gaps, indicating that the other DMUs are diverse in relation to $DMU_B$ and $DMU_H$. In rows R7 and R8, the analysis of $DMU_B$ and $DMU_H$ shows coordinates for points H and B above the prime meridian at (0.684, 1.503) and (0.494, 1.254), respectively.

The MCA technique, which integrates two-phase VGA-models for the identification of the optimal alternative, is effectively employed in the management of decision matrices distinguished by heterogeneous DMUs, thereby presenting a significant advantage over traditional MCA methodologies.

\subsection{Approaches to Address the MCA Challenge} \label{sec:8.2}
In Sections \ref{sec:4} and \ref{sec:5}, a thorough explanation of the two VGA-models is provided. MCA issues require consideration of both minimization and maximization criteria, as well as input and output metrics. The approaches are the same as \citet{Liu2025b}.

 \subsection{Accomplishments and Benefits of the New Approach} \label{sec:8.3}
An essential feature of MCA techniques is their capacity to assess multiple DMUs employing both \textit{quantitative and qualitative} performance indicators. However, existing MCA methods often suffer from subjective judgments and biases that compromise the dependability of the outcomes, even amidst considerable attempts to address these issues.

Sections \ref{sec:1.2} and \ref{sec:1.3} state that the limitations of the existing MCA methods can be addressed. However, our novel VGA-based MCA method objectively calculates \textit{unified goal prices} for $DMU_o$ in both Step I and Step II, eliminating the reliance on subjective judgment.
\subsection{Ranking DMUs} \label{sec:8.4} For an effective ranking of DMUs in MCA, the following steps should be followed. This novel method assesses each $DMU_o$ relative to others during both Stage I and Stage II, potentially identifying the worst alternative. Remove this alternative and repeat the process to progressively rank them as second, third, etc. If Stage II produces ties, such DMUs are deemed equally optimal. Additional analysis can be conducted if needed. Based on the information in Section \ref{sec:5}, the robustness sensitivity analysis can be bypassed if it is deemed non-essential.

\subsection{Extensions and Future Work on VGA-models} \label{sec:8.5}
\begin{enumerate}

\item In real-world decision-making scenarios, it might be essential to limit adjustment ratios. For example, by removing $\sum_{\forall i\in I} q_{io} \tau_o$ ($ \sum_{\forall r\in R}p_{ro} \tau_o$) from the goal functions in VGA-models, one can focus on either output-oriented or input-oriented assessments. In situations where \( X_a \) and \( Y_b \) remain unchanged, the objective functions across the eight VGA-models may be adapted to:

 $ \delta_o^\star = max \sum_{\forall i\in I\neq a} q_{io} \tau_o + \sum_{\forall r\in R\neq b}p_{ro}\tau_o$.
 
\item Typically, the adjustment ratios of inputs ($Q_o $) range from 0 to 1, while outputs ($P_o $) may exceed 1. The specific requirement can be satisfied by imposing constraints such as
($P_o  \leq 1$) or defining upper and lower bounds:
$\underline{Q}_o\le Q_o  \le \bar{Q}_o$ and $\underline{P}_o\le P_o  \le \bar{P}_o$.

\item The suggested approach is unable to handle decision matrices that include zero or negative values, posing a significant challenge in both research and practical applications.  

\item The MCDM and DEA literature has raised countless side conditions in the models. Adding side conditions to the VGA-models would enhance their applicability.  
\end{enumerate}
\section*{Acknowledgments}
This study was inspired by the DEA and MCDM methods. The authors thank the referees for providing their valuable feedback. 

\subsection*{Declaration of competing interest}
This research did not receive specific grants from public, commercial, or nonprofit funding agencies. The authors have no competing interests to disclose.

\subsection*{Data availability}
We confirm that the data supporting the findings of this study are available in this article.

\subsection*{CRediT authorship contribution statement}
\textbf{Su-Chuan Shih}: Conceptualization, Data collection, Validation, Formal Analysis, Visualization, Review and Editing, Supervision.
\textbf{Fuh-Hwa Franklin Liu}: Conceptualization, Methodology, and Writing of the Original Draft.

\medskip
\providecommand{\url}[1]{\texttt{#1}}
\providecommand{\urlprefix}{URL }

\end{document}